%% file: main.tex
\documentclass[acmsmall]{acmart}

\AtBeginDocument{%
  \providecommand\BibTeX{{%
    \normalfont B\kern-0.5em{\scshape i\kern-0.25em b}\kern-0.8em\TeX}}}

\setcopyright{acmlicensed}
\copyrightyear{2025}
\acmYear{2025}
\acmDOI{XXXXXXX.XXXXXXX}

\acmISBN{978-1-4503-XXXX-X/18/06}

\usepackage{framed}
\usepackage{tabularx}
\usepackage{graphicx}
\usepackage{multirow}
\usepackage{color, colortbl}

\newcommand{\control}{\textbf{\texttt{Control}}}
\newcommand{\program}{\textbf{\texttt{MST-GT}}}
\newcommand{\workflow}{\textbf{\texttt{MSTworkflow}}}
\newcommand{\workflowplus}{\textbf{\texttt{MSTworkflow+}}}

\newcommand{\revise}[1]{#1}
\newcommand{\ignore}[1]{}

\usepackage{xspace}
\newcommand{\etal}{\emph{et al.}\xspace}
\newcommand{\paratitle}[1]{\vspace{1.0ex}\noindent\textbf{#1}}
\newcommand{\ie}{\textit{i.e.,}~}
\newcommand{\eg}{\textit{e.g.,}~}

\begin{document}

\title{Fine-\revise{G}rained Appropriate Reliance: Human-AI Collaboration with \revise{a} Multi-\revise{S}tep {Transparent} Decision Workflow for Complex Task Decomposition}

\author{Gaole He}
\affiliation{%
  \institution{Delft University of Technology}
  \city{Delft}
  \country{The Netherlands}}
\email{g.he@tudelft.nl}

\author{Patrick Hemmer}
\affiliation{%
  \institution{Karlsruhe Institute of Technology}
  \city{Karlsruhe}
  \country{Germany}
}
\email{patrick.hemmer@kit.edu}

\author{Michael Vössing}
\affiliation{%
  \institution{Karlsruhe Institute of Technology}
  \city{Karlsruhe}
  \country{Germany}
}
\email{michael.voessing@kit.edu}

\author{Max Schemmer}
\affiliation{%
  \institution{Karlsruhe Institute of Technology}
  \city{Karlsruhe}
  \country{Germany}
}
\email{max.schemmer@kit.edu}

\author{Ujwal Gadiraju}
\affiliation{%
  \institution{Delft University of Technology}
  \city{Delft}
  \country{The Netherlands}
}
\email{u.k.gadiraju@tudelft.nl}
\renewcommand{\shorttitle}{Human-AI Collaboration with A Multi-step Transparent Decision Workflow}

\begin{abstract}
   In recent years, the rapid development of AI systems has brought about the benefits of intelligent services but also concerns about security and reliability. 
By fostering appropriate user reliance on an AI system, both complementary team performance and reduced human workload can 
be achieved. 
   Previous empirical studies have extensively analyzed the impact of factors ranging from task, system, and human behavior on user trust and appropriate reliance {in the context of one-\revise{step} decision making}. 
   However, user reliance on AI systems in tasks with complex semantics 
   that require multi-step workflows remains under-explored. 
   Inspired by recent work on task decomposition with large language models, we propose to investigate the impact of a novel {Multi-Step Transparent (MST) decision} workflow on user reliance behaviors. 
   We conducted an empirical study ($N=233$) of AI-assisted decision making in composite fact-checking tasks (\ie fact-checking tasks that entail multiple sub-fact verification steps). 
  Our findings demonstrate that human-AI collaboration with an {MST decision} workflow can outperform one-\revise{step} collaboration in specific contexts (\eg when advice from an AI system is misleading). 
  Further analysis of the appropriate reliance at fine-grained levels indicates that an MST decision workflow can be effective when users demonstrate a relatively high consideration of the intermediate steps. 
    Our work highlights that there is no one-size-fits-all decision workflow that can help obtain optimal human-AI collaboration. Our insights help deepen the understanding of the role of decision workflows in facilitating appropriate reliance. 
We synthesize important implications for designing effective means to facilitate appropriate reliance on AI systems in composite tasks\revise{, positioning opportunities for the human-centered AI and broader HCI communities.} 
\end{abstract}




\maketitle

\input{sections/sec-intro}
\input{sections/sec-related}
\input{sections/sec-task}
\input{sections/sec-study}
\input{sections/sec-exp}
\input{sections/sec-discussion}
\input{sections/sec-con}


\bibliographystyle{ACM-Reference-Format}
\bibliography{LLM_CoT_workflow}

\input{sections/sec-appendix}

\end{document}

%% file: sections/sec-intro.tex
\section{Introduction}

With the rapid development of artificial intelligence (AI) in recent years, 
there is a growing recognition of the promising value of AI assistance~\cite{amershi2019guidelines,bulten2021artificial}. 
AI systems have been used to answer knowledge-intensive questions~\cite{lan2022complex}, provide recommendations in e-commerce platforms~\cite{zhang2019deep}, and even help make critical decisions~\cite{sendak2020human}. 
While AI systems promise high effectiveness and use across domains, 
there is no guarantee of their correctness~\cite{li2023trustworthy,lai2021towards}. 
Thus, accountability and verifiability become a major concern before adopting such systems into existing workflows. 
To address these concerns, researchers and practitioners are actively exploring the potential of human-AI collaboration~\cite{lai2021towards,ehsan2021expanding,hemmer2021human,schemmer2022meta}. 
However, human-AI collaboration is not always effective, and 
there is a growing body of evidence suggesting that in many contexts human-AI team performance is inferior to AI performance alone~\cite{ehsan2021expanding,bansal2021does}. 
To address such issues and ensure complementary team performance {(\ie where the team performance can exceed the individual performance of both team members)}, users should accept advice from the AI system when it is correct and be able to override it when AI advice is incorrect. 
Such reliance patterns are denoted as appropriate reliance~\cite{schemmer2022should}, which has become a focal research topic at the intersection of AI and human-computer interaction.


In this context, existing work has explored how user trust and reliance are shaped by different aspects surrounding task characteristics~\cite{salimzadeh2023missing}, AI systems~\cite{Rechkemmer-CHI-2022}, and user factors~\cite{lai2021towards}. 
However, most of the research focuses on decision making or data annotation tasks which can be solved in a so-called \revise{\emph{one-step}} manner~\cite{salimzadeh2023missing}. 
In such a setting, a decision making task can be solved without requiring any intermediate steps. 
Herein human-AI collaboration allows humans to contrast their individual decisions against that of the AI, often enriched by further information, \eg its confidence or explanations on how a decision was derived~\cite{lai2021towards}. \revise{The ultimate goal here, is to enable humans to (hopefully) derive correct final decisions, leading to optimal team performance.} 
In contrast to the one-\revise{step} decision making setting, human-AI collaboration in complex multi-step decision making situations that require a composite semantic understanding and a multi-step workflow (\eg composite fact checking~\cite{aly2021feverous}) is still under-explored. 

In this work, we address this research gap by investigating the potential benefits and pitfalls of asking decision makers to follow the same multi-step workflow as that of an advisory AI system (\ie completing a sequence of decomposed sub-tasks) along with fine-grained transparency of AI systems. \revise{We consider the context of composite fact-checking due to its growing relevance in the age of LLMs, allowing us to simultaneously draw insights in a timely real-world task.}
The benefits of such a setup are two-fold. 
Firstly, such a workflow-based {decision process} enables us to analyze multi-step user decision making, where user decisions at the intermediate steps affect their final decision and reliance on the AI system. 
The key idea here is that following the same workflow \revise{as the AI system} can provide global transparency \revise{--- an overview of the process of the AI system (\ie task decomposition) ---} 
which allows users to check and verify intermediate steps of the AI system and better inform their reliance on AI advice. 
Secondly, each intermediate step can be viewed as a sub-task. 
Compared to global transparency, the sub-task information (\ie local decision criteria and evidence) entails local transparency (\ie at the level of a specific sub-task) of the intermediate decisions of the AI system. 
User reliance on the sub-task information (which is also input to the AI system)
provides a {fine-grained} view to analyze appropriate reliance on the AI system. 
{With the fine-grained transparency by design~\cite{mascharka2018transparency,felzmann2020towards}, we denote such \revise{a} multi-step workflow in our study as \textit{multi-step transparent (MST)} decision workflow.} 
{In this spirit, a multi-step transparent (MST) decision workflow can potentially facilitate appropriate reliance on the AI system and help advance our understanding of fine-grained user reliance.}

Although appropriate reliance has been extensively studied in relatively simple tasks~\cite{lai2021towards}, it is still unclear how user reliance is shaped by a {multi-step} decision workflow to solve complex tasks. 
When intermediate steps are adopted to solve complex tasks, users of the AI system may have more decisions to make sequentially. 
{For example, to verify the claim ``\textit{\revise{General Agreement on Trade in Services is a treaty created to extend the multilateral trading system to service sector and all members of the WTO are parties to the GATS.}}'', 
workers need to verify three sub-facts: 
\revise{(1) General Agreement on Trade in Services is a treaty;  
(2) General Agreement on Trade in Services is created to extend the multilateral trading system to service sector; 
(3) All members of the WTO are parties to the GATS.} 
In such a {multi-step} decision workflow, {accurate decisions at the intermediate steps} can be important. 
The intermediate steps and intermediate answers generated by the AI system can provide \textbf{global transparency} \revise{--- overall logic of the AI system (\ie complex fact-checking with task decomposition and answer aggregation).}
At the same time, the retrieved evidence at each intermediate step enables users to verify the intermediate answers generated by the AI system. 
In this way, it can increase the transparency of the AI system's intermediate decisions through verifiability~\cite{fok2023search}, which is denoted as \textbf{local transparency}. 
In this context, we propose to explore appropriate reliance on AI systems at the fine-grained level of intermediate steps and the level of task input of each step. 
In this paper, 
we address the following research questions: 
\begin{itemize}
    \item \textbf{RQ1}: How does a {multi-step} decision workflow shape user reliance on the AI system?
    \item \textbf{RQ2}: How do {global transparency and local transparency} shape user reliance in a {multi-step} decision workflow?
\end{itemize}

To this end, we conducted an empirical study ($N = 233$) in a composite fact-checking task (\ie identifying the factual accuracy of claims based on supporting documents). 
On the one hand, our findings provide empirical evidence that fine-grained appropriate reliance positively contributes to appropriate reliance at the level of overall task. 
With an {MST} decision workflow, users developed a fine-grained appropriate reliance on the intermediate steps, which enabled them to detect misleading AI advice.
On the other hand, {we found that an {MST} workflow does not}  improve human-AI team performance and appropriate reliance on AI advice in comparison to a \revise{one-step} decision workflow. 
In contrast to facilitating appropriate reliance globally, the {MST} decision workflow 
was effective only in a relatively challenging context, where AI advice is misleading. 
To encourage more precise intermediate decisions, we asked participants to reflect on the usefulness of supporting documents, which nudge users to carefully work on sub-tasks based on local transparency. 
We found such an intervention to increase user consideration in the intermediate steps brought about worse team performance and reliance patterns. 
{Combined with the cognitive load feedback across experimental conditions, we infer that such an intervention imposes a high cognitive load on users, limiting its expected impact. }
However, we found that the {MST} workflow can help users develop a critical mindset when making final decisions. 
This can partially explain why participants using an {MST} workflow showed decreased reliance on the AI system and their confidence decreased after access to the AI advice.

Our results highlight that the {multi-step transparent} decision workflow in complex tasks did have some positive impact in facilitating appropriate reliance. 
Appropriate reliance at the intermediate steps may be a prerequisite to making the {MST} decision workflow effective. 
While an {MST} workflow can help mitigate over-reliance \revise{in the presence of misleading AI advice,} 
it may also cause under-reliance without enough explicit considerations in the intermediate steps. 
We infer that there is no one-size-fits-all decision workflow to achieve optimal team performance in complex tasks. 
To this end, future work \revise{in the human-centered AI and relevant research communities} should explore how to dynamically adapt and combine multiple decision workflows according to the contextual requirements of human-AI collaboration. 
Our findings suggest that apart from the benefits of improving user consideration of the fine-grained transparency 
with specific interventions, it is important to consider potential trade-offs with concomitant side effects (\eg a high cognitive load caused by such interventions). 
{Finally, we identify promising future directions that explore how to improve human-AI team performance and promote global appropriate reliance by characterizing fine-grained appropriate reliance.} 
Our work has important theoretical implications for promoting appropriate reliance on AI systems in complex tasks and practical implications for the effective use of interventions to support human-AI collaboration. 

%% file: sections/sec-related.tex
\section{Related Work}

{Our work proposes to analyze fine-grained appropriate reliance on AI systems in handling complex tasks with a multi-step transparent decision workflow. Thus, we position our work in \revise{four realms} of related literature: \revise{trust calibration and} appropriate reliance in AI-assisted decision making (\ref{sec-AR}), multi-step hybrid workflows for complex tasks (\ref{sec-workflow}), transparency and verifiability of AI systems in human-AI collaboration (\ref{sec-rel-transparency})\revise{, misinformation and fact-checking (\ref{sec:rel-fact-checking})}.}


\subsection{\revise{Trust Calibration and} Appropriate Reliance in AI-assisted Decision Making}
\label{sec-AR}
Existing empirical studies~\cite{lai2021towards,bansal2021does,Lu-CHI-2021} and theoretical frameworks related to user trust~\cite{lee2004trust} and reliance behavior~\cite{schemmer2022should} highlight that users of an AI system need to identify when an AI system is accurate to rely on and when it is inaccurate and should be overridden. 
Such ideal reliance patterns are recognized as appropriate reliance on the AI system, but have proven to be 
extremely hard to obtain even by leveraging explainable AI methods~\cite{wang2021explanations}. 
\revise{Prior literature has adopted different definitions of trust; interpreting trust as either a subjective attitude or as objective user behavior in different contexts. 
Following the growing interpretation in AI-assisted decision making~\cite{lai2021towards,lee2004trust}, we operationalize user trust as a subjective attitude and user reliance as objective behavior in this work.} 
In most empirical studies~\cite{yin2019understanding,he2023stated,lai2021towards} where AI systems outperform human decision makers by a margin, the team performance has been reported to be typically worse than that of the AI alone. 
Addressing such challenges, empirical studies in one-\revise{step} decision making contexts have been proposed to mitigate under-reliance~\cite{wang2021explanations} (\ie disuse of accurate AI advice) and over-reliance~\cite{Chiang-IUI-2022,buccinca2021trust} (\ie misuse of misleading AI advice). 

\revise{Trust calibration has been extensively analyzed in interactions with AI systems~\cite{li2024developing,kaplan2023trust} and automation systems~\cite{lee1994trust,de2020towards,akash2020human,miller2021trust}. 
The primary goal is to align or adjust the level of trust that a human places in an AI system or automated technology based on the actual capabilities of that system. 
Prior work~\cite{miller2021trust,chen2018situation,mercado2016intelligent} has shown that transparency of the system (\eg pertaining to uncertainty or the reasoning process behind AI advice) can provide users with more situation awareness, and contribute to trust calibration. 
In particular, existing research has explored how information about AI performance~\cite{Rechkemmer-CHI-2022}, uncertainty of AI advice~\cite{tomsett2020rapid,kim2024m,salimzadeh2024dealing}, and reasoning process~\cite{vossing2022designing} affects user trust.
As pointed out by \citet{lee2004trust}, trust can substantially impact user reliance behaviors. 
Trust calibration has been shown to play an important role in facilitating appropriate reliance~\cite{kahr2024understanding}, aligning these lines of research. 
} 

Across multiple domains and diverse setups, researchers have found that many aspects surrounding user factors (like AI literacy~\cite{Chiang-IUI-2022} and cognitive bias~\cite{he2023knowing}), task characteristics (\eg task complexity~\cite{salimzadeh2023missing} and proxy task~\cite{buccinca2020proxy}), and AI transparency (\eg explainable AI~\cite{wang2021explanations}) have a substantial impact on user reliance. 
To mitigate the negative impact of these factors, researchers have proposed effective user interventions.
User tutorials 
have been proposed as an intervention that aims at educating users to fill in the knowledge gap~\cite{Lai-CHI-2020,Chiang-IUI-2022} and recognize the weaknesses of an AI system~\cite{chiang2021you}. 
Others have suggested performance feedback~\cite{he2023knowing,Rechkemmer-CHI-2022} through training sessions to calibrate user perceptions of the accuracy of an AI system. 
\citet{buccinca2021trust} proposed cognitive forcing functions to mitigate the illusion of explanatory depth~\cite{chromik2021think} brought about by explainable AI methods.



While existing work has explored how to evaluate and promote appropriate reliance on a global level, little is understood about user reliance behaviors on fine-grained levels in decision making contexts that go beyond one-\revise{step} decisions and require sequential decisions. 
In this work, we consider a composite fact-checking task as a test bed to explore how users leverage the intermediate steps and supporting documents in a {multi-step transparent} workflow. 
{Although multi-step workflows have been widely adopted in crowdsourcing~\cite{kittur2011crowdforge,kittur2012crowdweaver,gadiraju2014taxonomy} and crowd-AI hybrid systems~\cite{correia2023designing}, they have been under-explored in the context of AI-assisted decision making.}



\subsection{Multi-step Hybrid Workflows for Effective Task Completion}
\label{sec-workflow}


With the goal to obtain high-quality human annotations in complex tasks, prior {crowdsourcing} literature~\cite{kittur2011crowdforge,kittur2012crowdweaver,gadiraju2014taxonomy} has explored how to decompose complex tasks into multiple microtasks. 
To ensure text generation quality, Bernstein \etal~\cite{bernstein2010soylent} proposed the ``Find-Fix-Verify'' workflow, which splits complex text writing and editing tasks into a series of generation and review stages.
Through empirical studies on writing, brainstorming, and transcription, Little \etal~\cite{little2010exploring} 
found that both iteration and multiple votes can increase the average quality of responses, which is referred to as the ``Iterate-and-Vote'' workflow. 
With the rise of conversational agents in recent years, Qiu \etal~\cite{qiu2020improving} leveraged conversational microtask workflows to improve worker engagement. 
However, ~\citet{retelny2017no} argued that workflows can be a bottleneck to the effectiveness of crowdsourcing in complex tasks.

Inspired by such crowdsourcing literature, researchers have also proposed to build Crowd-AI Hybrid workflows~\cite{correia2023designing} to obtain high-quality data services.
For example, instead of obtaining fully manual annotations, asking crowd workers to follow a ``Find-Fix-Verify'' workflow may boost work efficiency and ensure high-quality outcomes~\cite{kamar2017complementing,wu2023llms}. 
Similar to the ``Iterate-and-Vote'' workflow, in a hybrid crowd-AI system, 
votes from crowd workers and AI systems can also improve outcome quality~\cite{zhang2019crowdlearn}. 
With the rise of large language models (LLMs), 
there is an increasing exploration of how conversational interaction can boost crowd-AI hybrid intelligence~\cite{shen2023convxai, slack2023explaining}. 
{For instance, users can obtain writing suggestions for a scientific paper using an LLM-powered conversational interface 
~\cite{shen2023convxai}. 
With a conversational human-AI interaction, users are involved in an implicit multi-step workflow to complete a task. 
Existing research has explored LLMs to automate exploratory conversations~\cite{He-IUI-2025} and plan daily tasks~\cite{he2025plan}.
Chaining multiple LLMs can achieve even complex functions entailed in music chatbots and writing assistants~\cite{wu2022promptchainer}. 
For example, Wu \etal~\cite{wu2022ai} defined primitive operations based on LLMs and chained them to synthesize controllable workflows dynamically. 
Such AI chains can also be adapted from crowdsourcing workflows~\cite{grunde2023designing}.}


We draw inspiration from existing literature on workflows for accomplishing tasks, and propose a multi-step transparent workflow for decision making in a complex fact-checking task. 
In our study, participants were required to go through intermediate steps of the AI \revise{(indicating the step-wise process of the AI)}, and verify the correctness of the final AI advice. 
\revise{Such a process allows users to develop an understanding of AI advice in a step-wise manner}, 
and 
make a final decision based on both AI advice and their initial decision. 
{The multi-step transparent workflow is generated and executed by the AI system~\cite{pan2023factchecking} (\ie LLMs coupled with retrieval-augmented generation\cite{lewis2020retrieval}) to provide advice and support participants in the task. 
Such human-AI collaboration increases the transparency of the AI system.} We aim to explore whether
 the increased transparency in such a process can facilitate appropriate reliance.


\subsection{Transparency and Verifiability in Human-AI Collaboration}
\label{sec-rel-transparency}

{Transparency has been recognized as an important goal towards building trustworthy AI systems~\cite{li2023trustworthy,kaur2022trustworthy,liao2023ai,ai2023artificial}. 
Existing work has explored the transparency of AI systems from different angles --- transparency in the reasoning process~\cite{kroll2021outlining}, transparency of data collection/curation~\cite{hutchinson2021towards}, transparency of limitations (\eg uncertainty)~\cite{smith2022real}, transparency of social context~\cite{green2020algorithmic,ehsan2021expanding} etc. 
Explainable AI (XAI) methods, which may be independent of the actual AI system, are also widely adopted to increase the transparency of AI systems in human-AI collaboration~\cite{ehsan2020human,liao2021human,ehsan2021expanding,wang2021explanations}. 
Besides incorporating XAI to increase system transparency, AI transparency is more explored theoretically~\cite{liao2023ai}. 
Relatively few works have attempted to empirically verify the impact of transparency on human-AI collaboration. 
With an empirical study, \citeauthor{vossing2022designing} found that providing the transparency of the reasoning process can increase user trust, while providing transparency of system uncertainty can decrease user trust~\cite{vossing2022designing}.}

{Different from the transparency of AI systems, verifiability is typically associated with specific AI advice. 
Within the context of human-AI collaboration, explainable AI methods~\cite{adadi2018peeking,saeed2023explainable} are widely used to assist human decision makers by providing evidence (\eg highlighting a part of task input~\cite{ribeiro2016should}) to support/oppose AI advice~\cite{wang2021explanations}. 
Among the explainable AI methods, causal explanations~\cite{mittelstadt2019explaining,chou2022counterfactuals} propose to reason about the causal relationships between the task input and AI advice, which provides a strong verifiability of AI advice. 
Recently, retrieval-augmented generation~\cite{lewis2020retrieval,fan2024survey} has emerged as one popular paradigm to enhance the verifiability of LLMs. 
With the retrieved evidence (\eg documents or relevant structure knowledge) as a reference, humans can verify the factual correctness of LLM generation.
}

{In this work, we followed the idea of transparency by design~\cite{mascharka2018transparency,felzmann2020towards} to modularize the complex fact-checking task into a series of sub-fact verification steps. 
With the decomposed sub-facts and sub-fact verification results, we provided users with global transparency of \revise{the AI system's overall process of task decomposition}. 
At the same time, we also provide the retrieved documents in each sub-fact verification, which are input to the LLM-based fact verification system. 
These documents provide local transparency of the intermediate steps (\ie sub-tasks). 
Thus, we provided fine-grained transparency of the AI system and explored how user reliance is shaped through the multi-step transparent (MST) decision workflow. 
To the best of our knowledge, this is the first empirical effort to understand user reliance on an AI system with fine-grained transparency.  

\subsection{\revise{Misinformation and Fact-checking}}
\label{sec:rel-fact-checking}
\revise{From a data mining perspective~\cite{shu2017fake}, misinformation is mainly detected based on two criteria: \textit{veracity} and \textit{intentionality}. 
Veracity mainly focuses on whether referred media or an online post is factually false or inaccurate, regardless of intent. 
`Fact-checking' is a task mainly based on veracity, which assesses whether claims made in written or spoken language are factually correct~\cite{thorne2018automated,guo2022survey,roozenbeek2024psychology}. 
Intentionality is another dimension based on the intent of the information creator/provider. For example, hate speech~\cite{mondal2017measurement} and `fake news' in the political election~\cite{guess2018selective}. 
Such misinformation, which often uses inflammatory
and sensational language to alter people's emotions~\cite{jiang2018linguistic}, can be harmful and widespread online~\cite{volkova2017separating,volkova2018misleading}. 
Based on these criteria, different communities have developed deep learning-based methods~\cite{guo2022survey,jahan2023systematic} to automate checking the massive amount of information online. In this work, we focus on the veracity of factual claims and conducted fact-checking tasks in a human-AI collaborative setting. }

\revise{While deep learning has been widely adopted to manage misinformation online, human partnership is still a crucial factor in this task~\cite{nguyen2018believe}. 
In addition to domain experts who are capable of detecting inaccurate or false information, researchers have explored and showcased crowdsourcing as an effective means to conduct fact-checking~\cite{nguyen2018interpretable,kim2018leveraging,arif2017closer,roitero2020can,saeed2022crowdsourced,allen2021scaling}. 
Typically, crowdsourced fact-checking involves three steps in a complex workflow~\cite{saeed2022crowdsourced}: (1) claim selection, which targets selecting check-worthy claims; (2) evidence retrieval, which obtains necessary information sources (\eg with a web browser); and (3) claim verification, which includes discussion and aggregation of judgment across different crowd workers and further produces explainable, convincing verdicts (\ie justification production). 
Prior to the rapid adoption of LLMs, the AI assistant in each stage of the complex workflow was typically trained independently and served different purposes. 
Such disparity between AI systems in different stages prevents humans from building a coherent and unified mental model when working with these sub-tasks. 
Recent advances have led researchers to explore leveraging LLMs to enhance all sub-tasks and provide an end-to-end workflow 
by chaining LLMs~\cite{pan2023factchecking}.}

\revise{It is evident that LLMs bring new opportunities and challenges to the fact-checking task~\cite{augenstein2024factuality,manakul-etal-2023-selfcheckgpt,chen2024combating}. 
On the one hand, LLMs have shown powerful natural language understanding and generation capabilities that can help tackle sub-tasks of fact-checking systems~\cite{li2024self}. 
For example, LLMs can retrieve highly relevant information sources~\cite{fan2024survey} and generate explanations to justify the verification process or the results~\cite{pan2023factchecking}. 
Furthermore, LLMs can provide an easy way for humans to communicate with the AI system, offering further potential 
for human-AI interaction in fact-checking tasks~\cite{augenstein2024factuality}. 
On the other hand, LLMs are known to hallucinate~\cite{ji2023survey}, \ie generating seemingly plausible but incoherent or factually incorrect content. 
LLMs have been shown to suffer from out-of-distribution data issues~\cite{yuan2023revisiting} and evolving knowledge without external contextual input (\eg retrieved documents)~\cite{fan2024survey}. 
Due to the uncertainty brought about by these prevalent flaws and the lack of accountability, human-AI collaborative fact-checking (comprising at least human oversight) is of fundamental importance in the era of LLMs.} 

\revise{In a user study of AI-assisted fact-checking, Nguyen~\etal~\cite{nguyen2018believe} found that crowd workers can be easily misled by wrong model predictions, but such errors can be reduced given interactions with the AI system. 
With dynamic user input and updated AI system predictions, crowd workers make much fewer errors misled by wrong AI predictions. 
Thus, Nguyen~\etal~\cite{nguyen2018believe} argued that `transparent models are key to facilitating effective human interaction with fallible AI models.' 
Contributing to existing literature in the area of human-AI collaboration for fact-checking, our work provides a multi-step transparent decision workflow in assisting humans conduct fact-checking with fine-grained retrieved evidence and decomposed sub-steps. Through this, we aim to provide fine-grained transparency and facilitate appropriate reliance of humans on the AI system. 
Our insights add further empirical evidence and advance our understanding of how transparency of the AI system and decision workflow affects human-AI interaction.
}

%% file: sections/sec-task.tex
\section{Task and Hypothesis}

In this section, we describe the composite fact-checking task (\ie identifying the factual accuracy of claims based on supporting documents), {the multi-step transparent workflow} (MST), and present our hypotheses, which have all been preregistered before any data collection.\footnote{\url{URL} hidden to preserve anonymity.} 


\begin{figure}[htbp]
    \centering
    \includegraphics[width=\textwidth]{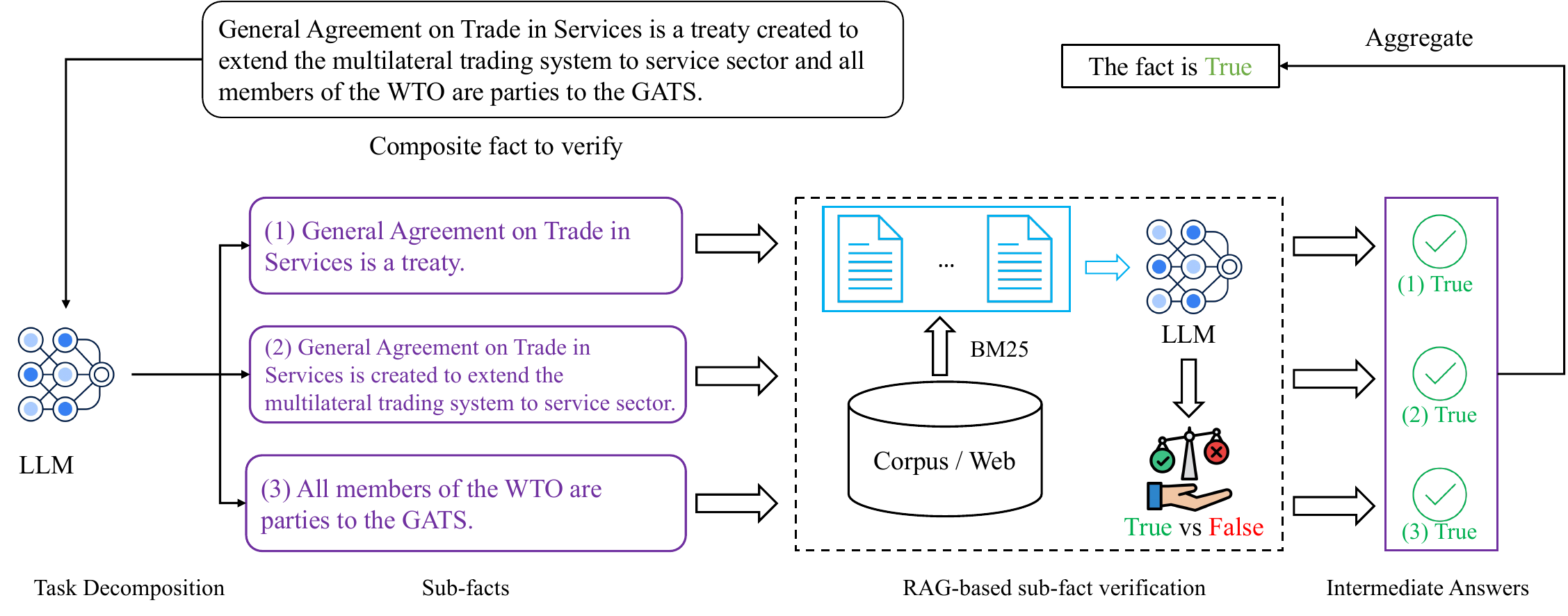}
    \caption{\revise{Illustration of the multi-step workflow on the composite fact-checking tasks using the ProgramFC method~\cite{pan2023factchecking}. The sub-facts and intermediate answers (in the purple box) provide global transparency in our MST workflow. The retrieved documents (in the blue box) serve as local transparency in our MST workflow.}}
    \label{fig:programfc}
    \Description{Illustration of the multi-step transparent workflow on the composite fact-checking tasks. The sub-facts and intermediate answers (in the purple box) provide global transparency in our MST workflow. The retrieved documents (in the blue box) serve as local transparency in our MST workflow.}
\end{figure}

\subsection{Composite Fact-checking Task}
\label{sec-task}
To analyze how the MST decision workflow impacts human-AI collaboration in complex tasks, we 
consider a composite fact-checking task. 
{An example of solving a composite fact-checking task based on the mutli-step transparent workflow is shown in Figure~\ref{fig:programfc}.} 
This task asks participants to decide whether a factual claim is \textbf{True} or \textbf{False} using the supporting documents retrieved from Wikipedia. 
The reasons for selecting the composite fact-checking task as our test bed are three-fold. 
Firstly, it contains tasks that require composite semantic understanding and can be solved with a workflow. 
Secondly, the fact-checking task requires evidence-based verification, which provides verifiability in the intermediate steps. 
Thirdly, due to the practical need for content moderation 
online (\eg hate speech, rumors, and hallucinated content from generative AI systems), it is \revise{a timely and relevant} scenario for human-AI collaboration. 

\subsection{Multi-step Transparent Workflow}


\paratitle{AI System Setup}. In our study, we adopted an LLM-based method called \textit{ProgramFC}~\cite{pan2023factchecking} to serve as our AI system. 
{Figure~\ref{fig:programfc} illustrates how \textit{ProgramFC} provides global transparency and local transparency.} 
The \textit{ProgramFC} method conducts fact-checking with two stages: (1) 
Using GPT-3.5 to generate decomposed steps to conduct composite fact-checking. 
(2) After generation of the decomposed steps, these steps are executed using another LLM, \textit{flan-t5-xl}~\cite{chung2022scaling}. Using the generated decomposed steps (\ie sub-facts to verify), the execution step generates intermediate answers based on retrieved supporting documents for each sub-fact. 
\revise{The documents are retrieved based on the popular BM25 algorithm~\cite{robertson1994some}, which leverages the query terms frequency appearing in documents to achieve a ranking function. All source documents are from Wikipedia, which is provided with the implementation of \textit{ProgramFC}.\footnote{\url{https://github.com/teacherpeterpan/ProgramFC}}}
Finally, \textit{ProgramFC} aggregates the intermediate answers to obtain a final prediction of the factual accuracy for the composite fact. 
The generated decomposed steps, intermediate answers, and retrieved supporting documents form the basis for the {multi-step transparent} workflow in our study. 
\revise{In our implementation, we selected the aforementioned LLMs due to two reasons: (1) \textit{flan-t5-xl} are representative open-sourced LLMs that are widely adopted in question answering and fact-checking practice~\cite{chung2024scaling}, (2) GPT-3.5 is representative of the performance of most open-sourced and commercial LLMs at the time of data collection (\ie Jan 2024), offering transferable findings and implications within the scope of our empirical study.}

\paratitle{Decision Workflow}. In our study, all workflows follow a two-stage decision making setup, a widely adopted design in AI-assisted decision making~\cite{wang2021explanations,Lu-CHI-2021,Chiang-IUI-2022,he2023knowing}. 
In the first stage, participants work on the fact-checking tasks based on the provided supporting documents {and the decision workflow}. 
Next, they were given a chance to alter their initial choice following AI advice. 
In multi-step decision workflows, decomposed steps and intermediate AI predictions are also shown to support user decisions. 
In the first stage of decision making, if participants do not find useful supporting documents to support or refute the sub-fact / fact, they can choose `Uncertain' beside the label `True' and `False'. 
In the second stage of decision making, participants are asked to make a binary decision between `True' and `False.'

\begin{figure}[h]
    \centering
    \includegraphics[width=\textwidth]{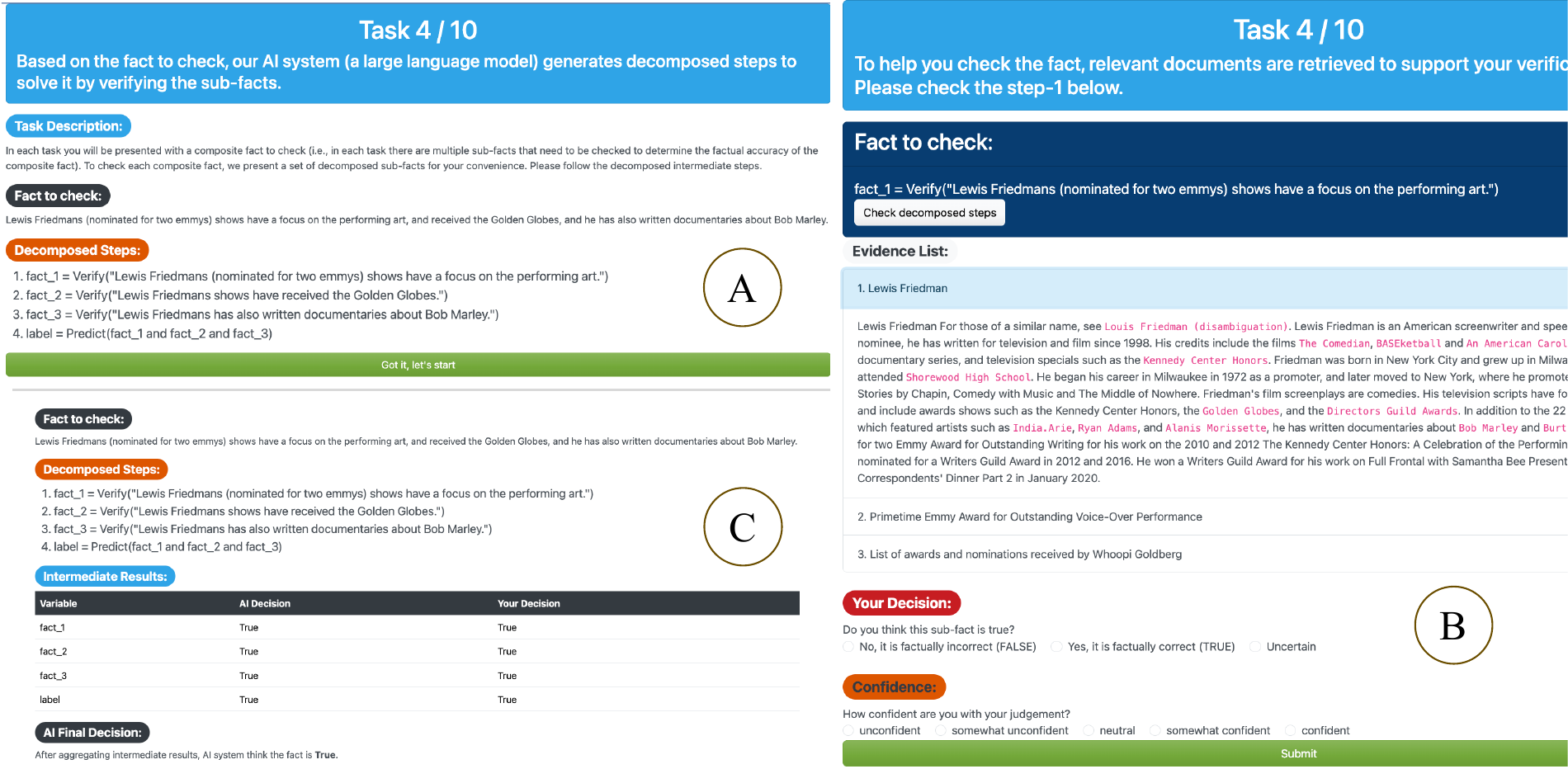}
    \caption{Screenshots of the composite fact-checking task interface with the MST workflow. (A) The starting point of the MST workflow, where the fact to check and decomposed steps are shown to users. (B) An intermediate step in the MST workflow. (C) Final decision making page, where decomposed steps and intermediate answers are provided as an explanation to the AI advice.}
    \label{fig:interface}
    \Description{Screenshot of our task interface. It consists of three sub-figures: (A) a page showing the decomposed steps, which is at the beginning of the MST workflow. (B) One intermediate step in MST workflow is from condition MSTworkflow. (C) The final decision page. On this page, decomposed steps and AI intermediate advice are provided as an explanation.}
\end{figure}

\paratitle{User Interface}. 
{The user interface of our study} is shown in Figure~\ref{fig:interface}. 
At the beginning of the {MST} workflow, we show participants the composite fact to check and decomposed steps (Figure~\ref{fig:interface} (A)). 
Next, participants are asked to follow the decomposed steps to verify the sub-facts (Figure~\ref{fig:interface} (B)) based on supporting documents. In our study, each step is provided with three relevant documents retrieved using the BM25 algorithm~\cite{robertson1994some}. 
Finally, participants receive the final AI advice on the factual accuracy of the composite fact. 
The decomposed steps and the corresponding AI advice at intermediate steps are also provided as explanations for the final AI advice. As a baseline comparison to the MST workflow, 
we also adopted a basic one-\revise{step} fact-checking workflow, where participants were asked to identify the factual accuracy of composite facts directly without explicitly decomposed steps. 
To ensure a fair comparison in terms of the evidence presented to participants across the two conditions, we gathered all documents retrieved in the three steps of the {MST} workflow and showed them on one page for the basic fact-checking workflow.

\subsection{Hypotheses}

Our study mainly aims to contribute to understanding user reliance behaviors on AI systems in the context of solving complex tasks in a sequence of decomposed sub-tasks \revise{generated by LLMs}. 
To this end, we devised a {multi-step} decision workflow and catered to {transparency} along the decomposed steps and AI advice at the intermediate steps. 
{The intermediate steps and answers work in facilitating a global transparency of the AI system, rendering the decision making process visible~\cite{vossing2022designing}.} 
Within our study, the \revise{decomposed steps generated by LLMs} show the \revise{overall step-wise process} of the AI system, which can potentially cause user trust to increase~\cite{vossing2022designing}. 
Meanwhile, user trust has been found to substantially impact user reliance behaviors~\cite{lee2004trust}. 
Combining these findings from existing work, we infer that providing \revise{decomposed steps generated by LLMs (\ie global transparency)} can increase user reliance on the AI system. 
Thus, we hypothesize that:
\begin{framed}
\textbf{(H1)} Compared to only providing final AI advice, providing \revise{the decomposed steps generated by LLMs (\ie global transparency)} will increase user reliance on the AI system.
\end{framed}

With the multi-step transparent workflow, users follow the same process \revise{(\ie decomposed steps in the same order)} to verify how the AI system works on the composite fact-checking task. 
{Throughout this process, the retrieved documents are provided to increase the transparency of each step (\ie sub-task). 
Based on local transparency, users make independent judgments about the intermediate steps before being exposed to \revise{final AI advice and intermediate answers predicted by the AI system.}
\revise{In comparison, users with a one-step decision workflow do not have the chance to work on the decomposed sub-tasks following the same step-wise process of the AI system.} 
Thus, \revise{users with a multi-step transparent workflow} may develop a more critical mindset when adopting the final AI advice supported with intermediate steps and answers.}
In this way, they may be better equipped to recognize when the AI system provides correct advice and when they should rely on their own decisions. 
Thus, we hypothesize that:

\begin{framed}
\textbf{(H2)} Providing users with a {multi-step transparent} decision workflow of the AI system will result in relatively more appropriate reliance on the \revise{AI system, {in comparison to a one-step decision workflow with AI advice}}. 
\end{framed}

{Although the intermediate steps and task input at each step (\eg retrieved documents) are designed to provide benefits in the decision making process, 
adequate consideration and appropriate use can be a prerequisite for their effectiveness.}
If users can properly leverage fine-grained transparency (\ie developing precise decisions and reflections at the level of intermediate steps and the level of task input in each step), 
they can benefit from the {multi-step} decision workflow, 
thereby calibrating their reliance behaviors on the AI system. Thus, we hypothesize that:
\begin{framed}
\textbf{(H3)}  Within a multi-step decision workflow, more accurate intermediate user decisions will result in relatively more appropriate reliance on the final AI advice.
\end{framed}


%% file: sections/sec-study.tex
\section{Study Design}
This study was approved by the human research ethics committee of our institution. 
Our hypotheses and experimental setup had all been preregistered before any data collection.

\subsection{Experimental Conditions}


Addressing the aforementioned RQs, we aim to explore the impact of a transparent decision workflow on user reliance on an AI system in a composite decision making task.  
Considering transparency of the decision workflow as the sole independent variable in our study, we designed a between-subjects study with four experimental conditions (see Table~\ref{tab:exp_condition}). 
In all conditions, participants follow a two-stage decision making setup (described in Section~\ref{sec-task}). The different experimental conditions are presented below, with each successive condition being a variant of the previous condition by a single factor.

\begin{enumerate}
    \item \control{} --- In this condition, participants follow a \revise{one-step} fact-checking workflow in the first stage and only have access to the final AI advice in the second stage.
    \item \program{} --- In this condition, participants can additionally check the intermediate steps from the AI system as global transparency in the second stage. 
    \item \workflow{} --- In this condition, participants follow a {multi-step transparent} workflow in the first stage, where they follow the same \revise{working logic (\ie decomposed steps in the same order)} of the AI system, and check the retrieved documents (\ie part of AI input) at each sub-task. In the second stage of decision making, they will be shown the intermediate steps and intermediate answers from both AI systems and themselves (cf. Figure~\ref{fig:interface}). 
    \item \workflowplus{} --- \revise{On top of condition \workflow{},} participants \revise{in this condition} are asked to annotate the usefulness of the supporting documents in each intermediate step. {Such annotation forces users to carefully check each retrieved document at the intermediate steps and indicate their usefulness in informing their intermediate decisions. This is designed to function similarly to cognitive forcing functions~\cite{buccinca2021trust}, which nudge users towards critical use of AI advice.}
\end{enumerate}
 



\begin{table}[htbp]
	\centering
	\caption{Differences between experimental conditions. The intermediate steps and answers are regarded as global transparency. Users have access to local transparency through the multi-step transparent workflow.}
	\label{tab:exp_condition}
    \begin{small}
	\begin{tabular}{c | c | c }
	    \hline
        \textbf{Exp Condition} & \textbf{Decision Workflow} & \textbf{AI Assistance}\\
        \hline
        \hline
        \control{}& \revise{one-step}  workflow& AI Advice\\
        \hline
        \program{}& \revise{one-step} workflow& AI advice + global transparency\\
        \hline
        \workflow{}& multi-step transparent workflow & AI advice + global transparency\\
        \hline
        \multirow{2}{*}{\workflowplus{}}& multi-step transparent workflow  & \multirow{2}{*}{AI advice + global transparency}\\
        & + document usefulness annotation& \\
    \hline
	\end{tabular}
 \end{small}
\end{table}

\subsection{Task Selection}
\revise{As described earlier, composite fact-checking is an important avenue for human-AI collaboration (\eg credibility assessment systems~\cite{robbemond2022understanding}). All data used in our study is from a public fact-checking dataset -- FEVEROUS-S~\cite{aly2021feverous}. 
This dataset is widely used in composite fact-checking, which leverages documents as evidence.}
The ten selected tasks in our study are shown in Table~\ref{tab:selected_tasks}.


\begin{table}[htbp]
	\centering
	\caption{\revise{Selected tasks in our study. `Extra Notes' provides special cases of the correctness of intermediate AI advice and verifiability of intermediate steps.}}
	\label{tab:selected_tasks}%
	\scalebox{.8}{
    \begin{tabular}{p{0.03\textwidth}|p{0.8\textwidth}|p{0.07\textwidth}|p{0.07\textwidth}|p{0.15\textwidth}}
		\hline
		\textbf{ID}& \revise{\textbf{Decomposed Steps generated by LLMs}}& \textbf{Ground Truth}& \textbf{System Advice}& \textbf{Extra Notes}\\
		\hline \hline
		1& \revise{(1) In 2014, both Orient Express and its holding company were renamed Belmond and Belmond Ltd, respectively. (2) Orient Express is a hospitality and leisure company that operates luxury hotels, train services and river cruises worldwide. (3) Belmond Ltd partnered with Irish Rail in 2015 to launch the luxury train Belmond Grand Hibernarian in Ireland.} & True& False& two misleading intermediate advice\\
        2& \revise{(1) Adrian Haynes is a Wampanoag chief. (2) Adrian Haynes served in the United States Navy during WWII from 1943 to 1947. (3) Adrian Haynes had a stint with the Naval Supply Ninth Amphibian Force that took part in the 1944 Anzio invasion in Italy.}& True& True& one intermediate step is not verifiable\\
        3& \revise{(1) Lewis Friedmans (nominated for two emmys) shows have a focus on the performing art. (2) Lewis Friedmans shows have received the Golden Globes. (3) Lewis Friedmans has also written documentaries about Bob Marley.} & True& True& - \\
        4& \revise{(1) General Agreement on Trade in Services is a treaty. (2) General Agreement on Trade in Services is created to extend the multilateral trading system to service sector. (3) All members of the WTO are parties to the GATS.} & True& True& - \\
        5& \revise{(1) Edgar McInnis wrote poetry in his spare time. (2) Edgar McInnis won the Newdigate Prize in 1925 for his poem "Byron". (3) Edgar McInnis received Master of Arts degree in 1930 from Oxford University.} & True& True& - \\
        \midrule
        6& \revise{(1) Teldenia strigosa was described by Warren in 1903. (2) Teldenia strigosa was found in New Guinea and Goodenough Island (in the Solomon Sea). (3) The length of the forewings of Teldenia strigosa is 12.5–15 mm.} & False& True& one misleading intermediate advice \\
        7& \revise{(1) The Travelling Church emigrants did not take any slaves with them when they traveled. (2) The Travelling Church emigrants traveled over the frozen and danger-filled Cumberland Gap. (3) The Cumberland Gap is a pass through the long ridge of the Cumberland Mountains, and within the Appalachian Mountains.} & False& True& one misleading intermediate advice\\
        8& \revise{(1) Tosi Fasinro finished fourth at the 1990 World Junior Championships. (2) Tosi Fasinro won the 1993 UK Championships. (3) Tosi Fasinro took one gold and one bronze at the AAA Championships.} & False& False& - \\
        9& \revise{(1) Stephanie Flanders was BBC's economics editor for five years. (2) Stephanie Flanders presented the docu-series Masters of Money. (3) Iain Duncan Smith praised Stephanie Flanders because of her pro-Labour stand in the coverage of unemployment figures.} & False& False& one intermediate step is not verifiable \\
        10& \revise{(1) The wild water buffalo or Asian buffalo is an endangered species. (2) The wild water buffalo or Asian buffalo is likely to become extinct shortly. (3) The wild water buffalo or Asian buffalo has a population of less than 1,000, of which the majority is found in India.} & False& False& - \\
		\hline
    \end{tabular}}
\end{table}%


\paratitle{Selection Process}. First, we generate decomposed steps for all tasks in the evaluation set of the FEVEROUS-S dataset. \revise{The task decomposition is achieved with prompting LLMs (GPT-3.5 in our study, but this can be easily replaced with other LLMs). The prompt is based on the implementation of ProgramFC~\cite{pan2023factchecking}.\footnote{\url{https://github.com/teacherpeterpan/ProgramFC/blob/main/models/prompts.py}}}
Next, we considered and retained all tasks that can be solved by verifying $3$ sub-facts. 
This resulted in 1,127 candidate composite tasks. 
The \textit{ProgramFC} algorithm achieved $67.7\%$ accuracy on these tasks. 
Estimating that each fact-checking task could take around 2-3 minutes for participants to complete, we selected ten tasks from these candidates. 
Considering all possible cases of (Ground Truth, AI Prediction) pairs, we randomly sample 10 tasks for each case (resulting in 40 tasks as candidates). 
An author of this paper then annotated the correctness of the decomposed steps and manually followed the decomposed steps to annotate both the usefulness of each supporting document and the factual accuracy of each sub-fact in the decomposed steps. 
After that, 15 tasks, where the decomposed steps were correct to verify the composite fact, were reserved. 

To balance the label distribution (True/False for the answer of each task), we selected five tasks with ground truth  ``True'' and five tasks with ground truth ``False'' (ten tasks intotal). 
$70\%$ accuracy is adopted when selecting the tasks, the rationale behind is: (1) it is very close to the actual AI accuracy $67.7\%$ (2) With such an accuracy level, the AI system is compatible with crowd workers to provide decision support without risking optimal performance with over-reliance, which makes it suitable to analyze user (appropriate) reliance patterns. 
To control the difficulty of tasks where AI prediction is wrong, tasks 1, 6, and 7 contain one or two incorrect intermediate steps. 
Besides, tasks 2 and 9 contain one intermediate step where the supporting documents are not enough to conclude the factual accuracy. 
In the two tasks, the AI final advice and the intermediate steps are all correct.



\subsection{Measures and Variables}

\subsubsection{Reliance-based dependent variables} 

In condition \workflow{} and \workflowplus{}, participants have to assess the intermediate correctness of each sub-fact. 
\revise{Each task is decomposed into three sub-tasks, which can facilitate valid comparison across conditions.}
Based on the user assessments of {the factual correctness of intermediate steps} and the ground truth (obtained through expert annotation), 
we can measure appropriate reliance at intermediate steps (\textbf{AR-Intermediate}) \revise{as average accuracy of user intermediate decisions}. 
Participants in condition \workflowplus{} were asked to annotate the usefulness of supporting documents when verifying each sub-fact with a question: ``\textit{Does this excerpt contain necessary information to verify the sub-fact?}''. 
There are four potential responses: ``\textit{Useless: it does not contain any useful information to verify the fact}'', ``\textit{Partial support: it contains some information partially support the sub-fact, but not fully support}'', ``\textit{Full support: it contains all necessary information to support the sub-fact}'', ``\textit{Contradiction: it contains necessary information to contradict with the sub-fact}''. 
To analyze how users appropriately leverage the intermediate supporting documents \revise{(\textbf{AR-Evidence})}, we adopted expert annotation of the usefulness of each supporting document as ground truth and calculated users' agreement ratio. 
\revise{Both \textbf{AR-Intermediate} and \textbf{AR-Evidence} are averaged across intermediate steps of the workflow.}


Since we aim to analyze the impact of different decision workflows on team performance and (appropriate) reliance, we leveraged measures introduced by prior work and typically used in this context due to their suitability
~\cite{yin2019understanding,Zhang-FAT-2020,schemmer2022should,he2023knowing}; we adopted \textit{Team Performance} (\ie average user accuracy based on their final decision) and \textit{Team Performance-wid} (\ie average user accuracy where their initial decision disagrees with AI advice) to measure user performance in our study. 
Following previous work by~\citet{yin2019understanding,Zhang-FAT-2020}, we measured user reliance by using the \textbf{Agreement Fraction} and the \textbf{Switch Fraction}. 
These measures consider the degree to which user final decisions agree with AI advice, and how often they switch to AI advice when their initial decision disagrees with the AI advice. 
Following prior work~\cite{schemmer2022should} on the evaluation of appropriate reliance, we adopted \textit{Relative positive AI reliance} (\textbf{RAIR}) and \textit{Relative positive self-reliance} (\textbf{RSR}) as appropriate reliance measures. {A low \textbf{RAIR} indicates under-reliance on the AI system, while a low \textbf{RSR} indicates over-reliance on the AI system.} \revise{To provide a precise definition of reliance-based measures used in our study, we provide further details of calculation formula in Appendix.}



\subsubsection{Other Variables} 
We also adopted measures for user trust, cognitive load, and relevant covariates to further our understanding of the impact of different decision workflows.

\paratitle{User Trust}. Motivated by existing work on user trust in automation~\cite{lee2004trust}, we assessed user subjective trust with a post-task questionnaire. 
We adopted four sub-scales from the trust in automation questionnaire~\cite{korber2019theoretical}: Reliability/Competence (TiA-R/C), Understanding/Predictability (TiA-U/P), Intention of Developers (IoD), and Trust in Automation (TiA-Trust).

\paratitle{Cognitive Load}. As the decision workflows in our study provide different levels of transparency of the AI system, there is a potential for the workflows to pose varying cognitive load among users. 
We assessed user cognitive load using the NASA-TLX questionnaire~\cite{colligan2015cognitive}. 

\paratitle{Covariates}. For a deeper analysis of our results, and to account for potential confounds based on exisiting literature, we  considered the following covariates: 
\begin{itemize}
    \item Familiarity with the AI system (\textit{Familiarity}) and general propensity to trust (\textit{Propensity to Trust}) from the trust in automation questionnaire.
    \item User expertise in large language models (\textit{LLM Expertise}) is assessed with a question ``To what extent are you familiar with large language models?''. Responses were gathered on a 5-point Likert scale from 1 (No prior experience/knowledge) to 5 (Extensive prior experience/knowledge).
    \item User expertise in fact-checking tasks (\textit{Fact Checking Expertise}) is assessed with a question ``Do you have any experience or knowledge with fact-checking?''. Responses were gathered on a 5-point Likert scale from 1 (No prior experience/knowledge) to 5 (Extensive prior experience/knowledge).
\end{itemize}

\begin{table*}[htbp]
	\centering
	\caption{The different variables considered in our experimental study. ``DV'' refers to the dependent variable. \textbf{RAIR}, \textbf{RSR}, and \textbf{Accuracy-wid} are indicators of appropriate reliance.}
	\label{tab:variables}
	\begin{footnotesize}
    \scalebox{.85}{
	\begin{tabular}{c | c | c | c}
	    \hline
	    \textbf{Variable Type}&	\textbf{Variable Name}& \textbf{Value Type}& \textbf{Value Scale}\\
	    \hline \hline

	    \hline
	    \multirow{2}{*}{Performance (DV)}& Team Performance& Continuous, Interval& [0.0, 1.0]\\
	    & Team Performance-wid& Continuous& [0.0, 1.0]\\
	    \hline
	    \multirow{2}{*}{Reliance (DV)}& Agreement Fraction& Continuous, Interval& [0.0, 1.0]\\
	    & Switch Fraction& Continuous& [0.0, 1.0]\\
        \hline
	    \multirow{4}{*}{Appropriate Reliance (DV)}& RAIR& Continuous& [0.0, 1.0]\\
	    & RSR& Continuous& [0.0, 1.0]\\
        & AR-Intermediate& Continuous& [0.0, 1.0]\\
        & AR-Evidence& Continuous& [0.0, 1.0]\\
	    \hline
     \multirow{4}{*}{Trust (DV)}& Reliability/Competence& Likert& 5-point, 1: poor, 5: very good\\
	    & Understanding/Predictability& Likert& 5-point, 1: poor, 5: very good\\
	    & Intention of Developers& Likert& 5-point, 1: poor, 5: very good\\
	    & Trust in Automation& Likert& 5-point, 1:strong distrust, 5: strong trust\\
     \hline
    \multirow{6}{*}{{Cognitive Load  (DV)}}& Mental Demand& Likert& -7: very low, 7: very high\\
	    & Physical Demand& Likert& -7: very low, 7: very high\\
	    & Temporal Demand& Likert& -7: very low, 7: very high\\
	    & Performance& Likert& -7: Perfect, 7: Failure\\
        & Effort& Likert& -7: very low, 7: very high\\
        & Frustration& Likert& -7: very low, 7: very high\\
     \hline
     \multirow{4}{*}{Covariates}& Propensity to Trust& Likert& 5-point, 1: tend to distrust, 5: tend to trust \\
     & Familiarity& Likert& 5-point, 1: unfamiliar, 5: very familiar \\
     & LLM Expertise& Likert& 5-point, 1: No expertise, 5: Extensive expertise \\
     & Fact Checking Expertise& Likert& 5-point, 1: No expertise, 5: Extensive expertise \\
     \hline
	    Other& Usefulness of Evidence& Category& \{useless, partial support, support, contradiction\}\\
	    \hline
	\end{tabular}}
	\end{footnotesize}
\end{table*}

Table~\ref{tab:variables} presents an overview of all the variables considered in our study.

\subsection{Participants}
\label{sec-participants}

\paratitle{Sample Size Estimation}. 
We computed the required sample
size in a power analysis for a Between-Subjects ANOVA using G*Power~\cite{faul2009statistical}. 
In our experimental analysis, we applied a Bonferroni correction to correct for testing multiple hypotheses. 
The significance threshold decreased to $\frac{0.05}{3}=0.017$, and is applied to all statistical analyses. 
We specified the default effect size $f = 0.25$, a significance threshold $\alpha = 0.017$ (\ie due to testing multiple hypotheses), a statistical power of $(1 - \beta) = 0.8$, and that we will investigate four different experimental conditions/groups. 
This resulted in a required sample size of $230$ participants. 
We thereby recruited 284 participants from the crowdsourcing platform Prolific,\footnote{\url{https://www.prolific.co}} accommodating potential exclusion.

\paratitle{Compensation}. 
To ensure a fair comparison across conditions, we set the basic payment for all participants to \pounds 4. This payment is based on the time estimation of condition \workflowplus{} (30 minutes) and a ``\textit{Fair}'' payment criteria (\pounds 8 per hour) by the platform. 
To motivate participants to reach correct decisions with their best ability, we rewarded each correct decision with a bonus of \pounds 0.05. 
Such a setup is also regarded as a contextual requirement to achieve appropriate trust in automation~\cite{lee2004trust}. 

\paratitle{Filter Criteria}. All participants were proficient English speakers aged between 18 and 50, and they had
finished more than 40 tasks and maintained an approval rate of at least 90\% on the Prolific platform.
We excluded participants from our analysis if they failed at least one attention check or if we found any missing data. 
In our study, frequently switching from initial agreement to opposite AI advice is treated as an indicator of potentially unreliable behavior. 
We excluded the participants with three or more such indications and participants who finished the study in a very short time (less than 15 minutes). 
These filter criteria ensure the quality of collected data by removing low-effort submissions. 
After these filter criteria, we have 233 participants reserved for analysis. These participants had an average age of 34 (SD = 7.5) and a reasonably balanced gender distribution (54.1\% male, 45.9\% female).

\subsection{Procedure}
\begin{figure}[htbp]
    \centering
    \includegraphics[width=\textwidth]{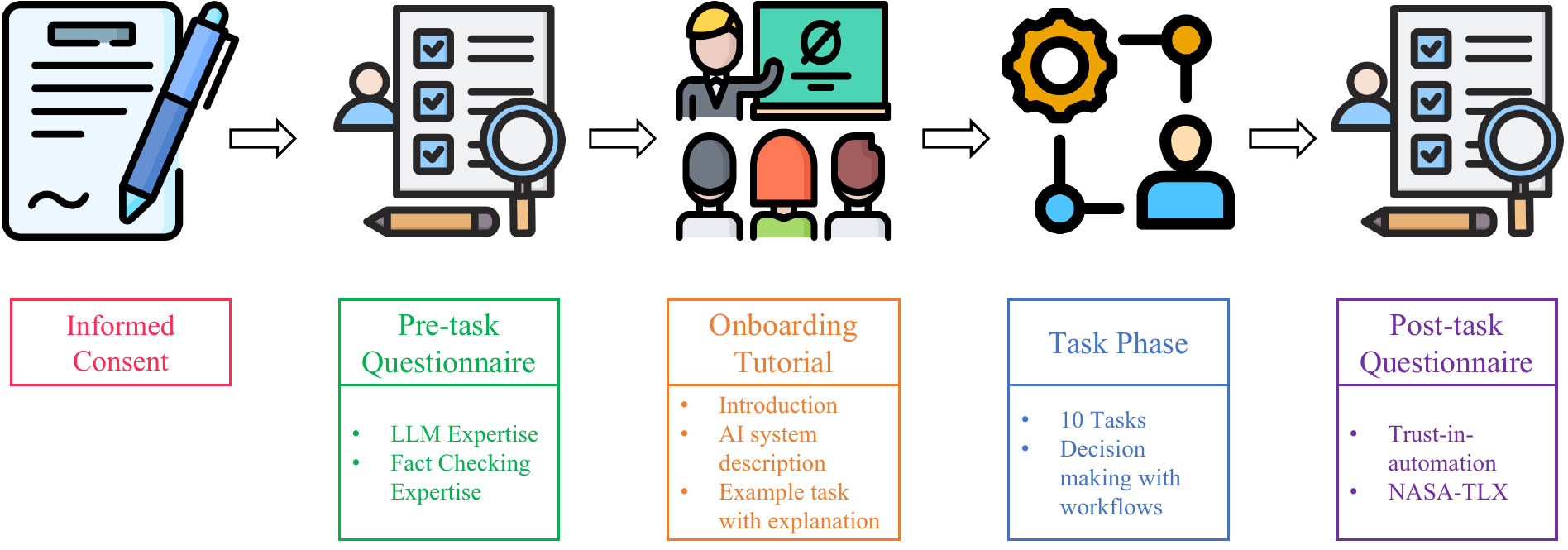}
    \caption{An illustration of the procedure that participants followed in our study.}
    \label{fig-procedure}
    \Description{Our study mainly contains five stages: (1) informed consent, (2) pre-task questionnaire, (3) onboarding tutorial, (4) task phase, (5) post-task questionnaire.}
\end{figure}

At the start of our study, all participants were asked to provide us with informed consent if they wished to proceed. 
Next, we gathered user self-reported expertise of large language models and fact-checking tasks using two questions. 
Before formally working on the tasks, we provide an onboarding tutorial to help participants get familiar with all elements shown in our study and understand how to work on the composite fact-checking tasks with an example. 
At this stage, participants in the condition \program{}, \workflow{}, and \workflowplus{} also have access to the decomposed steps and intermediate answers for the example task. 
Next, participants
worked on the ten selected tasks based on the decision workflow of the corresponding condition. 
Finally, they were asked to fill in post-task
questionnaires (including the trust in automation questionnaire and NASA-TLX questionnaire). 
We employed two attention check questions (one in the task phase and one in the post-task questionnaire) to ensure the quality of collected data.
Figure \ref{fig-procedure} illustrates the procedure participants followed in our study.

%% file: sections/sec-exp.tex
\section{Results} 
In this section, we will present the main experimental results and exploratory analysis for our study. 
In the spirit of open science, our code and data can be found in OSF repository.\footnote{\url{https://osf.io/s4he5/?view_only=2810286e1e5c4573b723c4785d5fe45c}}

\subsection{Descriptive Statistics}
To ensure the reliability of our results and interpretations, we only consider 233 participants who passed all attention checks. 
These participants were distributed across four experimental conditions in a reasonably balanced manner --- 54 (\control{}), 62 (\program{}), 60 (\workflow{}), 57 (\workflowplus{}).


\paratitle{Distribution of Covariates}. The covariates’ distribution is as follows: 
\textit{LLM Expertise} ($16.7\%$ with good/extensive background knowledge about LLMs, $83.3\%$ without any knowledge or with limited knowledge about LLMs), 
\textit{Fact Checking Expertise} ($28.8\%$ with good/extensive background knowledge with fact-checking tasks, $71.2\%$ without any knowledge or with limited background knowledge in fact-checking tasks), 
\textit{Propensity to Trust} ($M = 2.82$, $SD = 0.71$; 5-point Likert scale, \textit{1: tend to distrust}, \textit{5: tend to trust}), 
and \textit{Familiarity} ($M = 2.54$, $SD = 1.11$; 5-point Likert scale, \textit{1: unfamiliar}, \textit{5: very familiar}). 

\paratitle{Performance Overview}. On average across all conditions, participants achieved a \textit{Team Performance} of
$66\%$ ($SD = 0.12$), which is still lower than the AI accuracy ($70\%$). 
The average \textit{Agreement Fraction} is 0.78 ($SD = 0.14$), and the average \textit{Switch Fraction} is 0.50 ($SD = 0.30$). 
As for the appropriate reliance at fine-grained levels, participants in \workflow{} and \workflowplus{} conditions achieved an average \textit{AR-Intermediate} of 0.73 ($SD=0.12$); participants in \workflowplus{} condition achieved an average \textit{AR-Evidence} of 0.64 ($SD=0.24$). 
With these measures, we confirm that (1) participants in our study do not always blindly rely on AI advice, and (2) participants put some effort into the intermediate steps and supporting documents. 
As all behavior-based dependent variables (\ie performance, reliance, and appropriate reliance) are not normally distributed, we used non-parametric statistical tests to verify our hypotheses.

\begin{table}[htbp]
	\centering
	\caption{Participant performance on fact-checking tasks.  `Accuracy' is reported in percent (\%).  We use \textbf{bold} and \underline{underlined} fonts to denote the best and second-best performance in each task, respectively.
	}
	\label{tab:task-acc}%
        \begin{small}
        \scalebox{.95}{
	\begin{tabular}{c | c c | c c c c c}
	    \hline
	    \multirow{2}{*}{\textbf{Task-ID}}& \multirow{2}{*}{\textbf{System Advice}}& \multirow{2}{*}{\textbf{Ground Truth}}& \multicolumn{5}{c}{\textbf{Accuracy}}\\
        \cline{4-8}
         & & & \control{}& \program{}& \workflow{}& \workflowplus{}& Avg\\
	    \hline \hline
        1 & False & True & \textbf{33.3} & 15.0 & 14.5 & \underline{28.1} & 22.3\\
        \rowcolor{gray!15}2 & True & True & \textbf{79.6} & \underline{63.3} & 50.0 & 54.4 & 61.4\\
        3 & True & True & \textbf{70.4} & \underline{68.3} & 59.7 & 66.7 & 66.1\\
        \rowcolor{gray!15}4 & True & True & 94.4 & \textbf{100} & \underline{98.4} & 87.7 & 95.3\\
        5 & True & True & \underline{98.1} & \textbf{100} & 91.9 & 94.7 & 96.1\\
        \rowcolor{gray!15}6 & True & False & 9.3 & 8.3 & \textbf{25.8} & \underline{24.6} & 17.2\\
        7 & True & False & \underline{59.3} & 50.0 & \textbf{62.9} & 43.9 & 54.1\\
        \rowcolor{gray!15}8 & False & False & \underline{85.2} & \textbf{86.7} & 83.9 & 70.2 & 81.5\\
        9 & False & False & \underline{87.0} & \textbf{90.0} & 83.9 & 73.7 & 83.7\\
        \rowcolor{gray!15}10 & False & False & \underline{90.7} & \textbf{91.7} & 90.3 & 73.7 & 86.7\\
        
	    \hline
	\end{tabular}}%
    \end{small}
\end{table}%

Table~\ref{tab:task-acc} shows the accuracy across conditions in the ten selected fact-checking tasks. 
Among the seven tasks where system advice is aligned with the ground truth, participants in \control{} and \program{} conditions showed higher accuracy. 
However, participants in \workflow{} and \workflowplus{} conditions also showed competitive or even better performance in the three tasks where system advice was misleading.

\begin{figure}[htbp]
    \centering
    \includegraphics[width=\textwidth]{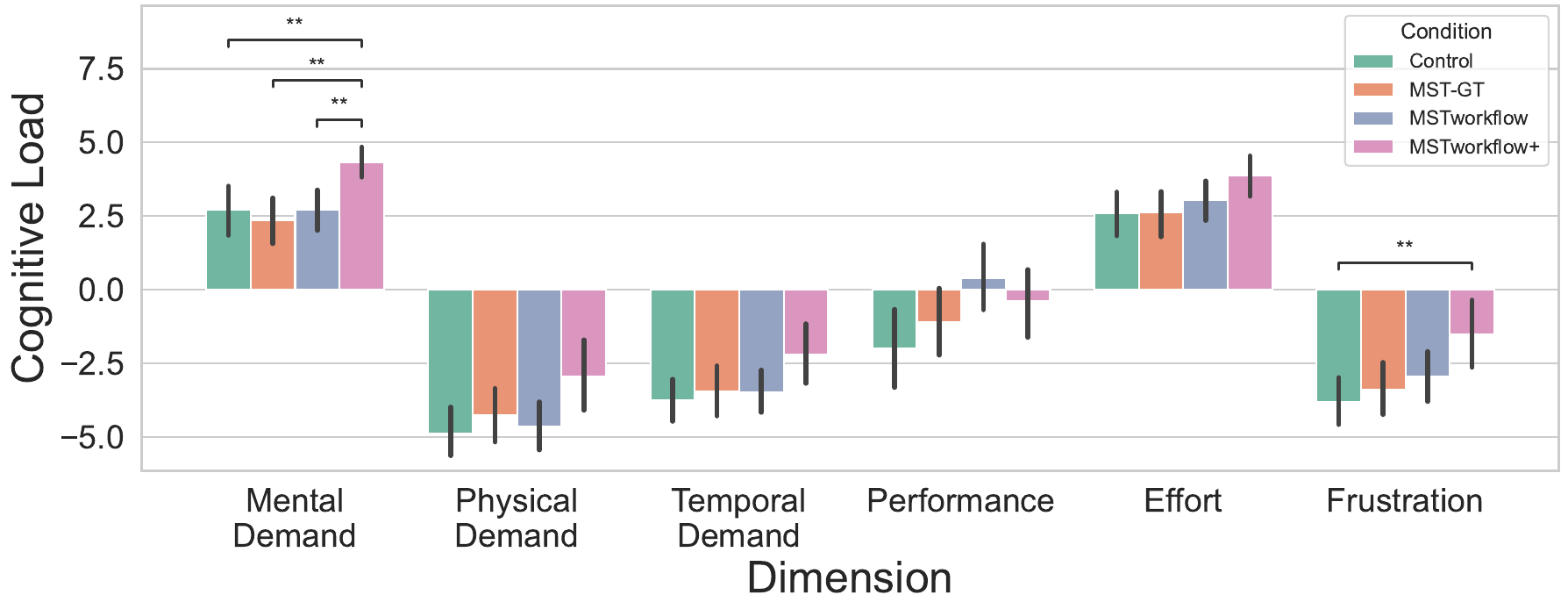}
    \caption{Bar plot illustrating the distribution of the cognitive load across different experimental conditions in our study. **: \textbf{\textit{p} < 0.017}}
    \label{fig-cognitive}
    \Description{A bar plot to illustrate the distribution of cognitive load across conditions in our study. Participants in condition MSTworkflow+ showed higher cognitive load in most dimensions. For mental demand and frustration, we found a significantly higher cognitive load. The significant pairs include: for mental demand, MSTworkflow+> Control, MST-GT, MSTworkflow; for frustration, MSTworkflow+ > Control.}
\end{figure}

\paratitle{Cognitive Load}. Based on the NASA-TLX questionnaire, we assessed participants' cognitive load based on six dimensions. 
As the workflows used in our study are of different levels of complexity and annotation effort. 
We visualized their distribution across conditions in Figure~\ref{fig-cognitive}. 
To compare the cognitive load across conditions, we conducted a one-way ANOVA test and post-hoc pairwise Tukey’s HSD test. 
The results indicate that participants in \workflowplus{} showed significantly higher \textit{Mental Load} than other conditions, and also showed much higher \textit{Frustration} when compared to the \control{} condition. While for the reserved four dimensions, the difference is non-significant, we still observe that participants in \workflowplus{} condition showed higher cognitive load.



\subsection{Hypothesis Tests}


\subsubsection{\textbf{H1} and \textbf{H2}: effect of different workflows}
To verify \textbf{H1} and \textbf{H2}, we conducted Kruskal-Wallis H-tests to compare the performance, reliance, and appropriate reliance measures of participants across the four experimental conditions. The results are shown in Table~\ref{tab:res-reliance}. 
For the measures where significant differences exist $p < 0.017$, we conducted a post-hoc Mann-Whitney test to obtain pairwise comparisons. 

\begin{table*}[hbpt]
	\centering
	\caption{Kruskal-Wallis H-test results for workflow on reliance-based dependent variables. ``${\dagger\dagger}$'' indicates the effect of the variable is significant at the level of 0.017.} 
	\label{tab:res-reliance}%
	\begin{scriptsize}
        \scalebox{0.82}{
	\begin{tabular}{c | c c | c c c c| c}
	    \hline
	    \multirow{2}{*}{\textbf{Dependent Variables}}& \multirow{2}{*}{$H$}& \multirow{2}{*}{$p$}& \multicolumn{4}{c|}{$M \pm SD$}& \multirow{2}{*}{\textbf{Post-hoc}}\\
     \cline{4-7}
     &  & & \control& \program& \workflow& \workflowplus\\
	    \hline \hline
	    \textbf{Team Performance}&15.13 & \textbf{.002}$^{\dagger\dagger}$& $0.71 \pm 0.11$ &$0.67 \pm 0.10$ &$0.66 \pm 0.13$ &$0.62 \pm 0.14$ & \control{} > \program{}, \workflow{} > \workflowplus{}\\
	\rowcolor{gray!15}\textbf{Agreement Fraction}& 16.37 & \textbf{.001}$^{\dagger\dagger}$& $0.80 \pm 0.11$ &$0.83 \pm 0.13$ &$0.75 \pm 0.13$ &$0.72 \pm 0.16$ &\control{}, \program{} > \workflow{}, \workflowplus{}\\
	\textbf{Switch Fraction}& 3.07 & .381& $0.47 \pm 0.27$ &$0.54 \pm 0.34$ &$0.46 \pm 0.26$ &$0.52 \pm 0.30$& -\\
    \rowcolor{gray!15}\textbf{Team Performance-wid}& 12.20 & \textbf{.007}$^{\dagger\dagger}$& $0.64 \pm 0.27$ &$0.52 \pm 0.26$ &$0.52 \pm 0.24$ &$0.49 \pm 0.21$& \control{} > \program{}, \workflow{}, \workflowplus{}\\
    \textbf{RAIR}& 3.64 & .302& $0.60 \pm 0.39$ &$0.54 \pm 0.42$ &$0.48 \pm 0.34$ &$0.52 \pm 0.33$& -\\
	\rowcolor{gray!15}\textbf{RSR}& 13.77 & \textbf{.003}$^{\dagger\dagger}$& $0.73 \pm 0.43$ &$0.47 \pm 0.47$ &$0.65 \pm 0.46$ &$0.46 \pm 0.48$& \control{}, \workflow{} > \program{}, \workflowplus{}\\
	    \hline
	\end{tabular}
        }
	\end{scriptsize}
\end{table*}


\paratitle{Impact on user reliance}. {As shown in Table~\ref{tab:res-reliance}, compared to  \control{} condition, \program{} condition showed significantly higher \textit{Agreement Fraction} and a non-significantly higher \textit{Switch Fraction}. This supports that providing transparency in the AI system's intermediate steps (\ie global transparency)  increases user reliance on the AI system. Thus, we find support for \textbf{H1}. 
However, with global transparency in final decision making, participants in conditions with a multi-step decision workflow (\ie \workflow{} and \workflowplus{}) showed significantly lower \textit{Agreement Fraction}.}

\paratitle{Impact on appropriate reliance}. {Although there is no significant difference on \textit{RAIR} across conditions. It is clear that participants in \control{} condition showed the highest \textit{RAIR} and \textit{RSR} corresponding to the highest level of appropriate reliance. 
Participants in \program{} and \workflowplus{} conditions showed significantly worse \textit{RSR} than \control{} and \workflow{} conditions, which is a reflection of over-reliance. 
Meanwhile, participants in the \workflow{} condition showed the worst \textit{RAIR}, which reflects a sub-optimal human-AI collaboration due to under-reliance. 
The reliance pattern differences between \program{} and \workflow{} conditions are a result of the decision workflow. 
Thus, we can infer that the {MST} decision workflow can help mitigate the over-reliance caused by providing global transparency. At the same time, it also introduces new issues of under-reliance. 
 Thus, we do not find support for \textbf{H2}. 

\subsubsection{\textbf{H3}: The impact of consideration in the intermediate steps}
As Table~\ref{tab:res-reliance} shows, participants in \workflowplus{} condition showed less reliance on the AI system and worst self-reliance (\textit{RSR} measure) and team performance compared to \workflow{} condition. 
In contrast to our expectation, the document usefulness annotation intervention fails to bring higher appropriate reliance at the intermediate steps (\ie \textit{AR-Intermediate}) -- \workflow{} condition achieved better \textit{AR-Intermediate} than \workflowplus{} condition. Thus, the results from Table~\ref{tab:res-reliance} are not sufficient to verify \textbf{H3}.

To analyze how explicit consideration in the intermediate steps shapes user reliance, we evenly re-split 
participants in \workflow{}  and \workflowplus{} conditions based on \textit{AR-Intermediate} 
-- \textbf{\texttt{{high consideration}}} ({with higher \textit{AR-intermediate}}, top $50\%$) and \textbf{\texttt{{low consideration}}} ({with lower \textit{AR-intermediate}}, bottom $50\%$).
Similar to the statistical analysis for \textbf{H1} and \textbf{H2}, we conducted Kruskal-Wallis H-test to compare the performance, reliance, and appropriate reliance measures of participants across the groups of participants. The results are shown in Table~\ref{tab:res-H3}. 

\begin{table*}[hbpt]
	\centering
	\caption{Kruskal-Wallis H-test results for \textbf{H3}. ``${\dagger\dagger}$'' indicates the effect of the variable is significant at the level of 0.017.}
	\label{tab:res-H3}%
	\begin{small}
        \scalebox{0.95}{
	\begin{tabular}{c | c c | c c| c}
	    \hline
	    \multirow{2}{*}{\textbf{Dependent Variables}}& \multirow{2}{*}{$H$}& \multirow{2}{*}{$p$}& \multicolumn{2}{c|}{$M \pm SD$}& \multirow{2}{*}{Post-hoc Results}\\
     \cline{4-5}
     &  & &  \textbf{\texttt{{high consideration}}}& \textbf{\texttt{{low consideration}}}\\
	    \hline \hline
	    \textbf{Team Performance}&21.90 & \textbf{.000}$^{\dagger\dagger}$& $0.70 \pm 0.11$ &$0.58 \pm 0.13$ & high > low\\
	\rowcolor{gray!15}\textbf{Agreement Fraction}& 7.43 & \textbf{.006}$^{\dagger\dagger}$& $0.78 \pm 0.14$ &$0.70 \pm 0.14$& high > low \\
	\textbf{Switch Fraction}& 2.42 & .120& $0.54 \pm 0.27$ &$0.44 \pm 0.28$& -\\
    \rowcolor{gray!15}\textbf{Team Performance-wid}& 7.97 & \textbf{.005}$^{\dagger\dagger}$& $0.56 \pm 0.22$ &$0.45 \pm 0.21$& high > low\\
    \textbf{RAIR}& 7.71 & \textbf{.005}$^{\dagger\dagger}$& $0.59 \pm 0.33$ &$0.41 \pm 0.31$& high > low\\
	\rowcolor{gray!15}\textbf{RSR}& 0.01 & 0.917& $0.56 \pm 0.48$ &$0.55 \pm 0.48$& -\\
	    \hline
	\end{tabular}
        }
	\end{small}
\end{table*}

For the measures we found a significant difference between the two groups, we conducted a post-hoc Mann-Whitney test to reach a conclusion.
As we can see, participants with higher explicit considerations of the intermediate steps in {MST} workflow will achieve higher \textit{RAIR} and a similar level of \textit{RSR}. 
It infers that low explicit considerations of the intermediate steps may cause under-reliance. 
Thus \textbf{H3} is supported by our experimental results.

We also found that participants with a high \textit{AR-Intermediate} achieved comparable performance across all conditions, which provides support for the effectiveness of the {MST} workflow. 
Meanwhile, participants in group \textbf{\texttt{low consideration}} showed much worse team performance, \textit{Agreement Fraction} and \textit{RAIR} (non-significant). 
This indicates that precise decisions at the intermediate steps play a critical role in making the MST workflow effective in human-AI collaboration.

\subsection{Exploratory Analysis}

\subsubsection{Impact of Trust}
Inspired by prior work~\cite{lee2004trust,he2023stated}, we analyzed how subjective user trust differs across conditions to provide further insights about the impact of different decision workflows. 
We conducted an \textit{Analysis of Covariance} (ANCOVA) with the \textit{decision workflow} as independent variable and \textit{TiA-Propensity to Trust}, \textit{TiA-Familiarity}, \textit{LLM Expertise}, and \textit{Fact Checking Expertise} as covariates. 
This allows us to explore the main effects of the decision workflow on subjective trust measured with the Trust in Automation questionnaire~\cite{korber2019theoretical}. Table~\ref{tab:trust} shows the ANCOVA results of the trust-related dependent variables. 
While there exists a borderline impact of decision workflow in \textit{TiA-R/C} and \textit{TiA-U/P}, the results are non-significant. 
Thus, we found that user trust in the AI system was not influenced by the decision workflow. 
However, we found that participants' general \textit{Propensity to Trust} had a significant impact on their trust. 
\textit{Propensity to Trust} shows a strong positive correlation with user trust, which will be detailed in Section~\ref{sec:correlation}.

\begin{table}[htbp]
	\centering
	\caption{ANCOVA test results on trust-related dependent variables. ``$\dagger$'' and ``$\dagger\dagger$'' indicate the effect of the variable is significant at the level of 0.05 and 0.017, respectively.
    } 
	\label{tab:trust}%
	\begin{footnotesize}
    \scalebox{.9}{
	\begin{tabular}{c | c c c | c c c| c c c | c c c}
	    \hline
	    \textbf{Dependent Variables}&	\multicolumn{3}{c|}{\textbf{TiA-R/C}}& \multicolumn{3}{c|}{\textbf{TiA-U/P}} & 
     \multicolumn{3}{c|}{\textbf{TiA-IoD}} & \multicolumn{3}{c}{\textbf{TiA-Trust}}\\
	    \hline
	    Variables& $F$& $p$& $\eta^2$& $F$& $p$& $\eta^2$& $F$& $p$& $\eta^2$& $F$& $p$& $\eta^2$\\
	    \hline \hline
	    Exp Condition&	2.78 & .042$^\dagger$ & 0.02 &3.31 & .021$^\dagger$ & 0.04 &1.63 & .183 & 0.01 &0.85 & .470 & 0.01 \\
        \rowcolor{gray!15}LLM Expertise& 2.33 & .128 & 0.01 &1.50 & .222 & 0.01 &1.00 & .319 & 0.00 &2.13 & .146 & 0.00\\
        Fact Checking Expertise& 1.83 & .178 & 0.00 &0.01 & .934 & 0.00 &0.17 & .676 & 0.00 &0.63 & .427 & 0.00\\
	    \rowcolor{gray!15}TiA-Propensity to Trust&	211.76 & \textbf{.000}$^{\dagger\dagger}$ & 0.47 &36.31 & \textbf{.000}$^{\dagger\dagger}$ & 0.13 &96.24 & \textbf{.000}$^{\dagger\dagger}$ & 0.29 &211.82 & \textbf{.000}$^{\dagger\dagger}$ & 0.48\\
	    TiA-Familiarity&	3.59 & .059 & 0.01 &0.01 & .922 & 0.00 &5.06 & .025$^\dagger$ & 0.02 &2.08 & .151 & 0.00\\
	    \hline
	\end{tabular}}%
	\end{footnotesize}
\end{table}%

\subsubsection{Impact of covariates} 
\label{sec:correlation}
In our study, we considered user expertise in large language models (\textit{LLM Expertise}), user expertise in fact-checking tasks (\textit{Fact Checking Expertise}), \textit{Familiarity}, and \textit{Propensity to Trust} obtained from questionnaires as covariates. 
These covariates may have a substantial impact on user trust and user reliance. 
To analyze how the covariates impact the dependent variables used in our study, we conducted Spearman rank-order tests between covariates and all dependent variables (\ie performance, reliance, appropriate reliance, trust, and cognitive load). 
The corresponding results are presented in Table~\ref{tab:correlation}. Our findings suggest that --- (1) Participants with a relatively higher \textit{LLM Expertise}, \textit{Fact Checking Expertise}, or \textit{Familiarity} 
reported a higher cognitive load. 
(2) Participants with a relatively higher \textit{LLM Expertise}, \textit{Familiarity}, or \textit{Propensity to Trust} also reported a higher level of trust in the AI system (TiA). 
(3) Participants with a relatively higher \textit{Propensity to Trust} exhibited a 
a higher \textit{Agreement Fraction} and \textit{Switch Fraction}, indicating more reliance on the AI system. However, the increased reliance may translate into over-reliance, as this corresponds with a significant negative correlation with \textit{RSR}. 
(4) Interestingly, higher \textit{LLM Expertise} does not necessarily help improve team performance in this task. Instead, the weak negative correlation between \textit{LLM Expertise} and \textit{Team Performance} suggests that participants with higher \textit{LLM Expertise} performed worse.

\begin{table*}[h]
	\centering
	\caption{Spearman rank-order correlation coefficient for covariates level on dependent variables. ``${\dagger\dagger}$'' indicates the effect of the variable is significant at the level of 0.017.}
	\label{tab:correlation}%
	\begin{small}
         \scalebox{0.92}{
	\begin{tabular}{l | c c | c c | c c | c c}
	    \hline
	    \textbf{Covariates}& \multicolumn{2}{c|}{LLM expertise} &  \multicolumn{2}{c|}{Fact checking expertise}& \multicolumn{2}{c|}{Familiarity}& \multicolumn{2}{c}{Propensity to Trust}\\
     \hline
     \textbf{Dependent Variables}&  $r$& $p$ &  $r$& $p$ &  $r$& $p$ &  $r$& $p$ \\
	    \hline \hline
    Mental demand & 0.073 & .270& 0.043 & .517& -0.094 & .152& 0.072 & .275\\
    Physical demand & 0.326 & \textbf{.000}$^{\dagger\dagger}$& 0.233 & \textbf{.000}$^{\dagger\dagger}$& 0.250 & \textbf{.000}$^{\dagger\dagger}$& 0.131 & .046\\
    Temporal demand & 0.158 & \textbf{.016}$^{\dagger\dagger}$& 0.093 & .157& 0.178 & \textbf{.007}$^{\dagger\dagger}$& 0.053 & .424\\
    Performance & -0.050 & .452& -0.104 & .112& -0.056 & .397& -0.020 & .763\\
    Effort & 0.152 & .020& 0.061 & .357& -0.072 & .272& 0.109 & .098\\
    Frustration & -0.020 & .757& -0.029 & .660& 0.017 & .795& -0.145 & .027\\
\hline
    Reliability/Competence & 0.215 & \textbf{.001}$^{\dagger\dagger}$& 0.035 & .597& 0.268 & \textbf{.000}$^{\dagger\dagger}$& 0.699 & \textbf{.000}$^{\dagger\dagger}$\\
    Understanding/Predictability & 0.146 & .026& 0.054 & .412& 0.106 & .106& 0.371 & \textbf{.000}$^{\dagger\dagger}$\\
    Intention of Developers & 0.272 & \textbf{.000}$^{\dagger\dagger}$& 0.138 & .035& 0.311 & \textbf{.000}$^{\dagger\dagger}$& 0.580 & \textbf{.000}$^{\dagger\dagger}$\\
    Trust in Automation (TiA) & 0.247 & \textbf{.000}$^{\dagger\dagger}$& 0.079 & .229& 0.273 & \textbf{.000}$^{\dagger\dagger}$& 0.725 & \textbf{.000}$^{\dagger\dagger}$\\
    \hline
    Team performance & -0.158 & \textbf{.016}$^{\dagger\dagger}$& -0.083 & .204& -0.143 & .029& -0.077 & .240\\
    Agreement Fraction & -0.017 & .791& -0.002 & .977& 0.006 & .924& 0.174 & \textbf{.008}$^{\dagger\dagger}$\\
    Switch Faction & 0.049 & .453& 0.036 & .582& 0.054 & .414& 0.217 & \textbf{.001}$^{\dagger\dagger}$\\
    Team Performance-wid & -0.127 & .052& -0.070 & .285& -0.149 & .023& -0.086 & .191\\
    RAIR & -0.046 & .480& -0.022 & .736& -0.015 & .823& 0.136 & .039\\
    RSR & -0.120 & .068& -0.087 & .187& -0.095 & .149& -0.235 & \textbf{.000}$^{\dagger\dagger}$\\
	    \hline
	\end{tabular}
        }
	\end{small}
\end{table*}


\subsubsection{Appropriate Reliance at Intermediate Steps} 
To further explore the relationship between the appropriate reliance in fine-grained levels (\ie \textit{AR-Intermediate} and \textit{AR-Evidence}) and global (appropriate) reliance, we conducted the Spearman rank-order test separately for participants in \workflow{} and \workflowplus{} conditions.
The results are shown in Table~\ref{tab:correlation-AR}. 
We found a strong positive monotonic relationship between the fine-grained appropriate reliance (\ie \textit{AR-Intermediate} and \textit{AR-Evidence}) 
and performance-based measures (\textit{Team Performance} and \textit{Team Performance-wid}). 
\textit{AR-Intermediate} also shows some positive impact on \textit{Agreement Fraction} and \textit{RAIR}, which indicates that participants with higher \textit{AR-Intermediate} have less chance to be impacted by the under-reliance issues. 
In contrast to what one can intuitively expect, we found that \textit{AR-Evidence} does not necessarily show a significant positive correlation with appropriate reliance. We will further discuss these findings in section~\ref{sec:discussion}.

\begin{table*}[hbpt]
	\centering
	\caption{Spearman rank-order correlation coefficient for AR-Intermediate and AR-Evidence on dependent variables. ``${\dagger\dagger}$'' indicates the effect of the variable is significant at the level of 0.017.}
	\label{tab:correlation-AR}%
	\begin{small}
        \scalebox{0.73}{
	\begin{tabular}{l | c c | c c | c c | c c | c c | c c}
	    \hline
	    \textbf{Dependent Variables}& \multicolumn{2}{c|}{Team performance} &  \multicolumn{2}{c|}{Agreement fraction}&  \multicolumn{2}{c|}{Switch faction}&  \multicolumn{2}{c|}{Team performance-wid}&  \multicolumn{2}{c|}{RAIR}&  \multicolumn{2}{c}{RSR}\\
     \hline
     \textbf{Fine-grained AR}&  $r$& $p$ &  $r$& $p$&  $r$& $p$ &  $r$& $p$&  $r$& $p$ &  $r$& $p$  \\
	    \hline \hline
    AR-Intermediate & 0.477 & \textbf{.000}$^{\dagger\dagger}$& 0.300 & \textbf{.000}$^{\dagger\dagger}$& 0.114& .217& 0.308& \textbf{.001}$^{\dagger\dagger}$& 0.270& \textbf{.003}$^{\dagger\dagger}$& 0.089&.337\\
    AR-Evidence & 0.474 & \textbf{.000}$^{\dagger\dagger}$& 0.227 & .090& -0.066& .624& 0.300& .023& 0.132& .329& 0.123& .362\\
	    \hline
	\end{tabular}
        }
	\end{small}
\end{table*}

\subsubsection{Confidence Dynamics}
User confidence in their decision also plays an important role in shaping their reliance on AI systems. We illustrated user confidence dynamics using a line plot (Figure~\ref{fig:confidence_dynamics}) based on their average confidence along with task order. For participants in \workflow{} and \workflowplus{} conditions, we calculated the average confidence (\ie Initial-avg) and minimum confidence (\ie Initial-min) of the three intermediate steps for their initial decision confidence.
Overall, participants were confident with both their initial and final decisions, which is around 4.0 (corresponding to ``\textit{somewhat confident}''). 
It is evident that the gap in confidence for participants in \control{} and \program{} conditions is relatively smaller than the participants in \workflow{} and \workflowplus{} conditions. 
We also found that the gap between the average initial confidence and minimum initial confidence is relatively stable (around 0.5 on a 5 point scale).
{Compared with \workflow{} condition, the participants in \workflowplus{} condition showed relatively lower initial confidence and final confidence. 
This reflects that participants indicated more uncertainty about their decisions in the presence of the document usefulness annotation in \workflowplus{} condition.

\begin{figure}[h]
    \centering
    \includegraphics[width=\textwidth]{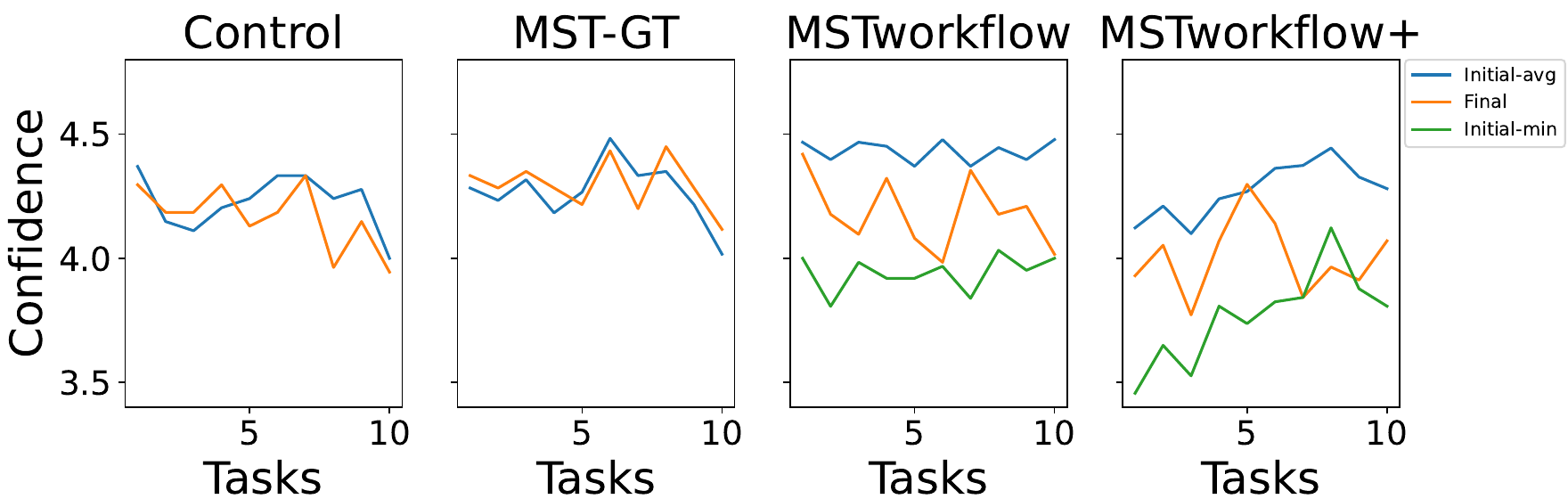}
    \caption{Line plot illustrating the confidence dynamics among users after receiving the AI advice (and explanations). The orange line and blue line illustrate the confidence dynamics before and after receiving AI advice (and explanations), respectively.}
    \label{fig:confidence_dynamics}
    \Description{Four subfigures illustrate the initial user confidence and final user confidence across four experimental conditions. For conditions CoTworkflow and CoTworkflow+, we aggregated initial user confidence with both minimum and average aggregators.}
\end{figure}

%% file: sections/sec-discussion.tex
\section{Discussion}
\label{sec:discussion}

\subsection{Key Findings}
Our experimental results show that the {multi-step transparent} workflow {can} be effective in specific contexts (\eg in challenging tasks where AI advice is misleading) and that appropriate reliance at intermediate steps 
is important to ensure effective human-AI collaboration and team performance. 
Our analysis of appropriate reliance at the intermediate steps highlights 
{that the MST workflow can facilitate human-AI collaboration when users make explicit considerations of the intermediate steps (cf. Table~\ref{tab:res-H3}).}
We also found that participants who do not demonstrate fine-grained appropriate reliance at the intermediate steps may exhibit under-reliance behavior on the final AI advice. 
Based on the user confidence dynamics, we found that participants with the {MST} workflow reported higher confidence in their initial decision, but also showed a decreased confidence after checking the AI advice and in the presence of {transparency cues such as intermediate steps and intermediate answers}. 
These findings can be explained as follows---participants with a {MST} workflow obtained more verifiability of the AI advice, which contributes to developing higher self-confidence (reflected by initial confidence) and a critical mindset on AI advice. 
These benefits may help mitigate over-reliance caused by the potential illusion of explanatory depth shown in condition \program{}, but they may also decrease user reliance on the AI system and cause under-reliance \revise{when they make errors in the intermediate steps.}

\paratitle{Impact of the Multi-step Transparent (MST) Workflow on User Reliance}. In our study, we found that global transparency (\ie \revise{overall logic of the AI system --- task decomposition and intermediate answers}) can have the negative impact of increasing over-reliance on the AI system (cf. Table~\ref{tab:task-acc} and Table~\ref{tab:res-reliance}). 
This is consistent with findings in explainable AI literature --- explainable AI can cause over-reliance on the AI system~\cite{buccinca2021trust,schemmer2022influence}. 
Although the {MST} workflow does not always work as expected to facilitate appropriate reliance on the AI system, our results suggest that it works well in some specific contexts (\eg challenging tasks where AI advice can be misleading). 
Similar findings have been pointed out in crowdsourcing literature -- crowdsourcing workflows can be useful in specific contexts but may also constrain complex work~\cite{retelny2017no}. 
{While the multi-step workflow can help address over-reliance on the AI system, it can also introduce under-reliance when users make mistakes at the intermediate steps. 
Such impact is highly similar to the impact of second opinion in AI-assisted decision making~\cite{lu2024does}. 
This is because, when there is no performance feedback, users may doubt the accuracy level of the AI assistance after disagreements with AI advice (caused by misleading second opinion / intermediate decision). 
As a result, users may decrease their trust and reduce reliance on the AI system.
}

\paratitle{Document Usefulness Annotation Did Not Work as Expected}. The cognitive forcing function of document usefulness annotation (condition \workflowplus{}) did not show the effectiveness to increase \textit{AR-Intermediate}. 
In contrast to our expectations, the intervention to increase user consideration of the evidence on top of the {MST} workflow resulted in decreased team performance and relatively lower appropriate reliance. 
{With such an intervention, users demonstrated less confidence in themselves at intermediate steps and final decisions. 
Consistent with our expectations, we found that participants in the condition \workflowplus{} reported much higher cognitive load than other conditions (cf. Figure~\ref{fig-cognitive}).} 
{A potential explanation is that the byproduct of a high cognitive load overrides the benefits of the multi-step transparent workflow. 
The decreased self-reported confidence in decisions can be a signal of uncertainty. As a result, users may turn to rely more on the AI system, which helps explain the over-reliance on misleading AI advice (cf. Table~\ref{tab:res-reliance}).} 

Our analysis of the impact of covariates (Section~\ref{sec:correlation}) also provides interesting insights: 
(1) participants who reported a relatively higher LLM expertise tended to show higher subjective trust but performed worse (cf. Table~\ref{tab:correlation});
(2) participants with a higher propensity to trust showed higher subjective trust, and higher reliance, in fact resulting in over-reliance. \revise{(3) Different from other covariates, the fact checking expertise does not show a strong correlation with trust and reliance on the AI system. 
Based on these observations, we can infer some user factors (\ie LLM expertise, familiarity, and propensity to trust in our study) can potentially increase user trust. However, the increased user trust is not calibrated according to the actual AI performance, which hinders effective human-AI collaboration.}
In summary, these user factors may lead to uncalibrated trust in the AI system, which causes over-reliance. These findings are consistent with previous empirical studies of AI-assisted decision making~\cite{he2023knowing,he2023stated}, and our work sheds light on how they extend to the context of human-AI decision making using a {multi-step transparent} workflow.
%

This is the first work that has explored how a {multi-step transparent} workflow shapes user reliance and when such a workflow can be effective. 
While previous work has extensively studied user reliance and appropriate reliance at a global level~\cite{lai2021towards}, 
our work is the first to explore the fine-grained levels of appropriate reliance --- appropriate reliance in the intermediate steps aligned with evidence. 
Our results indicate that participants who made better use of the intermediate steps and evidence achieved better team performance (cf. Table~\ref{tab:res-H3} and Table~\ref{tab:correlation-AR}). 
Their reliance patterns were also positively impacted by their consideration of the intermediate steps (\ie positive correlation between \textit{AR-Intermediate} and \textit{RAIR} in Table~\ref{tab:correlation-AR}). 
These findings suggest promising future directions to explore in the context of decision making with \textit{RAG}-based AI systems and human-AI collaboration with decision workflows.

\subsection{Implications}


\paratitle{Suitable workflows \revise{and user interventions} can ensure effective human-AI collaboration}. Our work has important implications for designing effective human-AI collaboration workflows.
In our study, we found that participants who followed a basic one-\revise{step} decision workflow performed better on relatively easy tasks. 
In comparison, participants who adopted a {multi-step transparent} workflow performed better on challenging tasks where AI advice was misleading. 
Our findings suggest that there is no one-size-fits-all solution for human-AI collaboration workflows. This echoes findings in previous analyses of workflows in crowdsourcing~\cite{retelny2017no}.
Multiple aspects in task characteristics (\eg task complexity~\cite{salimzadeh2023missing}), user factors (\eg cognitive bias~\cite{nourani2021anchoring,he2023knowing}), and system transparency (\eg reasoning process~\cite{mascharka2018transparency}) may impact the final decision outcome. 
As opposed to seeking and designing for optimal human-AI workflows that can always lead to high effectiveness, future work can explore how to combine multiple human-AI workflows depending on different contextual requirements. 
For example, in tasks where an AI system demonstrates low confidence, we may expect more useful and independent input from human decision makers. 
We would then need to adopt a suitable workflow (such as an {MST} workflow) to improve team performance, 
a critical mindset for critical consideration of AI advice and the verifiability of AI advice. 

We also found that some interventions (\ie condition \workflowplus{}) can pose a high cognitive load on users of the AI system, which can result in the side effects of user frustration and decreased effectiveness. 
This is consistent with prior findings of unforeseen negative impacts of user interventions~\cite{lai2021towards,bertrand2022cognitive}. 
Such a phenomenon has also been observed in prior empirical studies~\cite{schemmer2022influence,wang2021explanations,nourani2021anchoring,he2023knowing,lu2024does} of AI-assisted decision making. 
For example, some prior work~\cite{schemmer2022influence,wang2021explanations,nourani2021anchoring} found that explainable AI can help address under-reliance, but also simultaneously led to a higher over-reliance. Providing a second opinion shows a similar impact~\cite{lu2024does} on mitigating over-reliance while increasing under-reliance. 
These observations reveal that user interventions can be effective only in specific contexts. 
Therefore, a trade-off between the benefits and harms of user interventions can be a prerequisite for their effectiveness. 
Identifying the target audience and contextual requirements for user interventions based on behavioral and psychological patterns can be a promising direction to explore.

\paratitle{Fine-grained Analysis to Promote Appropriate Reliance}. 
{Our experimental results suggest that appropriate reliance at a global level and complementary team performance may be dependent on more fine-grained appropriate reliance on the intermediate steps (\ie global transparency) and supporting documents (\ie local transparency). 
While most existing work has explored a \textit{one-\revise{step}} decision workflow, the user decision making process and user decision criteria are not accessible for analysis. 
Existing empirical studies typically set up experiments with several conditions by controlling factors about user, task, and AI system. 
In such a setup, the user reliance on more fine-grained task input and the surrounding context (\eg relevant documents) are typically not considered for analysis. 
We argue that this limits us, as a community, from developing an insightful understanding of appropriate reliance on AI advice. 
In this spirit, our work has important methodological implications 
for both studying and promoting appropriate reliance with a fine-grained analysis.
\revise{Our findings and implications can help develop human-centered AI systems for complex tasks highlighting accountability, like medical diagnosis, loan prediction, supply chain optimization, etc. 
The users would benefit from the critical mindset and intermediate results of a multi-step transparent workflow.}
\revise{In the future, human-centered AI studies exploring appropriate reliance 
could consider more structured workflows for decision making and operationalize fine-grained user reliance.} 


\subsection{Caveats and Limitations}


\paratitle{Transferability Concerns}. {In our study, we used the lens of a single complex task --- composite fact-checking supported with retrieved documents, and a specific AI system --- an LLM-based system that first decomposes a complex fact and then verifies sub-facts by leveraging retrieval-augmented generation. 
It is unclear how our findings and implications can transfer to a different context where task characteristics (\eg difficulty, uncertainty and risks) and system characteristics (\eg transparency and system accuracy) are different~\cite{salimzadeh2023missing,salimzadeh2024dealing}. 
\revise{
It is noteworthy that in multi-step decision workflows, there can be dependencies between sub-tasks. In our study, we considered sub-tasks that are largely independent (\ie sub-facts could be treated independently). This presents a limited view of multi-step decision workflows and future research is needed to extend our work to workflows with dependencies. }
{Our setup of fine-grained transparency and multi-step decision workflow offers a relatively general framing to analyze multi-step human-AI collaboration and fine-grained appropriate reliance.} 
Future work can follow this framing to explore how different aspects surrounding task characteristics,
AI systems and user factors affect user trust and reliance in a multi-step human-AI collaboration.
}

\paratitle{Impact of cognitive load}. 
We found significant differences in the perceived cognitive load of participants across different experimental conditions, suggesting that the \workflowplus{} condition required participants to consistently exert relatively more effort. 
A qualitative follow-up or an in-person user study with a selection of participants may provide deeper insights. 
More work is needed to understand the effectiveness of such workflows that are demanding of users during the decision making process itself. 
Such workflows may be less suited in low-stakes compared to relatively high-stakes contexts, where cognitive effort can be considered as a viable trade-off for better team performance. 

\paratitle{Potential Bias}. 
As our study is carried out with crowd workers, participants with a {MST} workflow may spend more effort and feel more temporal demand. Their performance may be impacted by \textit{self-interest bias}~\cite{draws2021checklist}. 
To ensure the collected data is of high quality, we made sure that participants in each condition received a fair payment according to platform standards and provide bonuses to motivate correct decisions. 


%% file: sections/sec-con.tex
\section{Conclusion}
In this work, we conducted an empirical study to analyze how a {multi-step transparent (MST)} decision workflow shapes user reliance in a composite fact-checking task. 
Compared to a basic workflow and one-\revise{step} human-AI collaboration, participants with an {MST} workflow showed a more critical mindset in making decisions while relying on AI advice. 
As a result, the {MST} workflow tackled over-reliance on AI advice to some extent but also led to a decrease in user reliance on the final AI advice and user confidence, which may cause under-reliance (\textbf{RQ1}). 
When participants demonstrate explicit considerations of AI advice in the intermediate steps (\ie when they can achieve high appropriate reliance at fine-grained levels), such under-reliance can be mitigated. 
Then the {MST} workflow can 
facilitate effective human-AI collaboration. 
Increasing the transparency of the AI system {by providing intermediate steps and answers}
caused over-reliance on the AI system. 
At the same time, we found that the {MST} workflow with an additional task that attempts to cognitively engage participants by annotating the supporting documents and increase critical reflection may pose a demanding cognitive load on participants. 
This resulted in harming the overall human-AI collaboration, as reflected by the user experience, team performance, and appropriate reliance of participants in that experimental condition. 
Having said that, through further analysis (cf. Table~\ref{tab:res-H3}), we found 
appropriate reliance at the level of intermediate steps may be required to ensure the effectiveness of the {MST} workflow.
We also found that appropriate reliance on the retrieved evidence is positively correlated with team performance. Based on this finding, we infer that the transparency of the AI system at the level of task input can also play a positive role in facilitating {appropriate reliance and complementary} human-AI collaboration (\textbf{RQ2}). 
More work is required to further advance our understanding of this problem. 

Our results indicate that the {MST} workflow can be effective in specific contexts, and there is no one-size-fits-all decision workflow to achieve optimal human-AI collaboration. 
A trade-off between the benefits (\eg {fine-grained transparency} and critical consideration of AI advice, more verifiability) and side effects (\eg higher cognitive load) of decision workflows should be considered in human-AI collaboration. 
Our findings have important
implications for designing effective decision workflows to facilitate appropriate reliance and better human-AI collaboration.

%% file: sections/sec-appendix.tex
\appendix

\section{Appendix}  

\begin{table}[h]
	\centering
	\caption{\revise{Variable notations and their corresponding description.}}
	\label{tab:symbols}
	\begin{tabular}{l  l }
	    \hline
        \textbf{Variable Notation} & \textbf{Definition}\\
        \hline
        $M$ & Number of intermediate steps\\
        $N$ & Number of tasks\\
        $T$ & Task set, $T = \{t_1, t_2, ..., t_N\}$\\
        $t_{ij}$ & Sub-fact verification task (\ie task $t_i$, step j)\\
        $S(t)$ & Supporting documents used for task $t$\\
        $D(u, T_{ij})$ & Decision maker $u$'s decision for task sub-task $t_{ij}$\\
        $D_G(t)$ & Ground truth for task $t$\\
        $U(u, s)$ & Decision maker $u$'s  annotation of usefulness of supporting document $s$\\
        $U_E(s)$ & Experts' annotation of usefulness of supporting document $s$\\
    \hline
	\end{tabular}
\end{table}

\revise{To provide a precise definition of \textbf{AR-Intermediate} and \textbf{AR-Evidence}, we first denote all the related variables in Table~\ref{tab:symbols}. 
In our study, the tasks are sampled from the task set $T = \{T_1, T_2, ..., T_N\}$, where $N$ is the size of the task set. For each fact-verification task, we follow $M$ steps to conduct sub-fact checking. In our study, $N = 10$, and $M$ is fixed to 3 for every task to facilitate valid comparison across conditions. For the sub-fact verification, we can use $T_{ij}, i \in [1, N], j \in [1, M]$ to denote sub-fact verification for task $T_i$ at step $j$. 
Based on the user assessments of {the factual correctness of intermediate steps} and the ground truth (obtained through expert annotation), 
we can measure appropriate reliance at intermediate steps (\textbf{AR-Intermediate}).} 

\revise{$$\textnormal{\textbf{AR-Intermediate}} (u) = \frac{1}{N}\sum_{i=1...N}\frac{1}{M}\sum_{j=1...M}\mathbb{I} \left( (D(u, T_{ij}) = D_G(T_{ij}) \right).$$}

\revise{Participants in condition \workflowplus{} were asked to annotate the usefulness of supporting documents when verifying each sub-fact. 
To analyze how users appropriately leverage the intermediate supporting documents, we adopted expert annotation of the usefulness of each supporting document as ground truth and calculated users' agreement ratio. 
This can be achieved by comparing user ($u$) annotation of the usefulness of document $s$ ($U(u,s)$) with expert annotation of the same document ($U_E(s)$). All these annotations are task-specific. Specifically,}
\revise{$$\textnormal{\textbf{AR-Evidence}} (u) = \frac{1}{N}\sum_{i=1...N}\frac{1}{M}\sum_{j=1...M}\frac{\sum_{s \in S(T_{ij})}\mathbb{I}\left( (U(u, s) = U_E(s) \right)}{|S(T_{ij})|}.$$}

\revise{Following prior work~\cite{yin2019understanding,Zhang-FAT-2020,schemmer2022should,he2023knowing}, we also adopted other measures to assess performance (\textit{Team Performance} and \textit{Team Performance-wid}), reliance (\textit{Agreement Fraction} and \textit{Switch Fraction}), and appropriate reliance (\textit{RAIR} and \textit{RSR}). For the calculation of these measures, please refer to~\cite{schemmer2022should,he2023knowing}.}
